\documentclass[journal]{IEEEtran}
\usepackage{amsfonts,amsmath}
\usepackage{diagbox}
\usepackage{hyperref}
\usepackage{graphicx,subfigure}
\usepackage{booktabs}
\usepackage{algorithm}
\usepackage[noend]{algpseudocode}
\usepackage[noadjust]{cite}
\usepackage{xcolor}
\newcommand{\edit}[1]{#1}
\usepackage{tikz}
\usepackage{textcomp}
\usepackage{hyperref}
\usepackage{lipsum}

\newcommand\copyrighttext{%
  \footnotesize \textcopyright 2022 IEEE. Personal use of this material is permitted.  Permission from IEEE must be obtained for all other uses, in any current or future media, including reprinting/republishing this material for advertising or promotional purposes, creating new collective works, for resale or redistribution to servers or lists, or reuse of any copyrighted component of this work in other works. DOI: \href{https://ieeexplore.ieee.org/abstract/document/9765650}{10.1109/TASE.2022.3168621}}
\newcommand\copyrightnotice{%
\begin{tikzpicture}[remember picture,overlay]
\node[anchor=south,yshift=10pt] at (current page.south) {\fbox{\parbox{\dimexpr\textwidth-\fboxsep-\fboxrule\relax}{\copyrighttext}}};
\end{tikzpicture}%
}

\def\cA{{\mathcal A}}

\def\cM{{\mathcal M}}

\def\cO{{\mathcal O}}

\def\cS{{\mathcal S}}

\def\bbE{{\mathbb E}}

\def\bbR{{\mathbb R}}

\def\bbZ{{\mathbb Z}}

\begin{document}

\title{Unified Automatic Control of Vehicular Systems with Reinforcement Learning}

\author{Zhongxia~Yan,~\IEEEmembership{Member,~IEEE,}
        Abdul~Rahman~Kreidieh,~\IEEEmembership{Member,~IEEE,}
        Eugene~Vinitsky,~\IEEEmembership{Member,~IEEE,}
        Alexandre~M.~Bayen,~\IEEEmembership{Senior Member,~IEEE,}
        and~Cathy~Wu,~\IEEEmembership{Member,~IEEE}% <-this % stops a space
\thanks{Manuscript received January 7, 2022; revised March 10, 2022; accepted April 16, 2022. This article was recommended for publication by Associate Editor M. Robba and Editor J. Yi upon evaluation of the reviewers' comments. This work was supported in part by Amazon, in part by the MIT-IBM Watson AI Laboratory, and in part by the Department of Transportation Dwight David Eisenhower Transportation Fellowship Program. \textit{(Corresponding author: Zhongxia Yan.)}}
\thanks{Zhongxia Yan is with the Department
of Electrical Engineering and Computer Science, Massachusetts Institute of Technology, Cambridge,
MA, 02139 USA (e-mail: zxyan@mit.edu).}% <-this % stops a space
\thanks{Abdul Rahman Kreidieh is with the Department of Civil and Environmental Engineering, University of California, Berkeley, Berkeley, CA 94720 USA (e-mail: aboudy@berkeley.edu)}% <-this % stops a space
\thanks{Eugene Vinitsky is with the Department of Mechanical Engineering, University of California, Berkeley, Berkeley, CA 94720 USA (e-mail: evinitsky@berkeley.edu)}% <-this % stops a space
\thanks{Alexandre M. Bayen is with the Department of Electrical Engineering and Computer Science, University of California, Berkeley, Berkeley, CA 94720 USA (e-mail: bayen@berkeley.edu)}% <-this % stops a space
\thanks{Cathy Wu is with the Laboratory for Information and Decision Systems, the Department
of Civil and Environmental Engineering, and the Institute for Data, Systems, and Society, Massachusetts Institute of Technology, Cambridge,
MA, 02139 USA (e-mail: cathywu@mit.edu).}% <-this % stops a space
}

\markboth{IEEE Transactions on Automation Science and Engineering,~Vol.~, No.~, Month~2022}%
{Shell \MakeLowercase{\textit{et al.}}: Bare Demo of IEEEtran.cls for Journals}

\maketitle

\begin{abstract}
Emerging vehicular systems with increasing proportions of automated components present opportunities for optimal control to mitigate congestion and increase efficiency. There has been a recent interest in applying deep reinforcement learning (DRL) to these nonlinear dynamical systems for the automatic design of effective control strategies. Despite conceptual advantages of DRL being model-free, studies typically nonetheless rely on training setups that are painstakingly specialized to specific vehicular systems. This is a key challenge to efficient analysis of diverse vehicular and mobility systems. To this end, this article contributes a streamlined methodology for vehicular microsimulation and discovers high performance control strategies with minimal manual design. A variable-agent, multi-task approach is presented for optimization of vehicular Partially Observed Markov Decision Processes. The methodology is experimentally validated on mixed autonomy traffic systems, where fractions of vehicles are automated; empirical improvement, typically 15-60\% over a human driving baseline, is observed in all configurations of six diverse open or closed traffic systems. The study reveals numerous emergent behaviors resembling wave mitigation, traffic signaling, and ramp metering. Finally, the emergent behaviors are analyzed to produce interpretable control strategies, which are validated against the learned control strategies.
\end{abstract}

Note to Practitioners:
\begin{abstract}
As vehicular systems such as real-world traffic systems and robotic warehouses become increasingly automated, optimizing vehicle movements sees an increasing potential to reduce congestion and increase efficiency. For many vehicular systems, simulations of varying fidelity are commonly used for analysis and optimization without the need to deploy real vehicles. This article describes a unified and practical approach for optimal control of vehicles in arbitrary simulated vehicular systems while permitting partial automation, where the behavior of fractions of vehicles at given times can be modelled but not controlled. As illustrated by the diverse traffic systems considered in this article, the presented methodology emphasizes ease of application within any simulated vehicular system while minimizing manual efforts by the practitioner. The control inputs consist of local information around each automated vehicle, while the control outputs are commands for longitudinal acceleration and lateral lane change. Experimental results are presented for relatively small simulated traffic systems, though the methodology can be adapted to larger vehicular systems with minor modifications. Experimentally optimized behaviors provide insights to the practitioner which may assist in designing simplified and interpretable control strategies. Implementation in real-world systems depends on two requirements: 1) a reliable fallback mechanism for ensuring safety of vehicles, and 2) sufficient fidelity of the simulator for simulated behaviors to transfer. These requirements are under active research for traffic systems and may be practical in some robotic settings. \edit{To facilitate robust transfer of policies from simulated to real-world systems, future extensions of this work may inject additional randomization into simulation while reducing the unmodeled stochasticity of targeted real-world systems as much as possible.}
\end{abstract}

Primary and Secondary Keywords
\begin{IEEEkeywords}
Primary Topics: mobile traffic control, automated vehicles, reinforcement learning
Secondary Topic Keywords: mixed autonomy, multi-agent systems
\end{IEEEkeywords}

\IEEEpeerreviewmaketitle
\copyrightnotice

\begin{figure*}[!hbt]
\centering
\includegraphics[width=\textwidth]{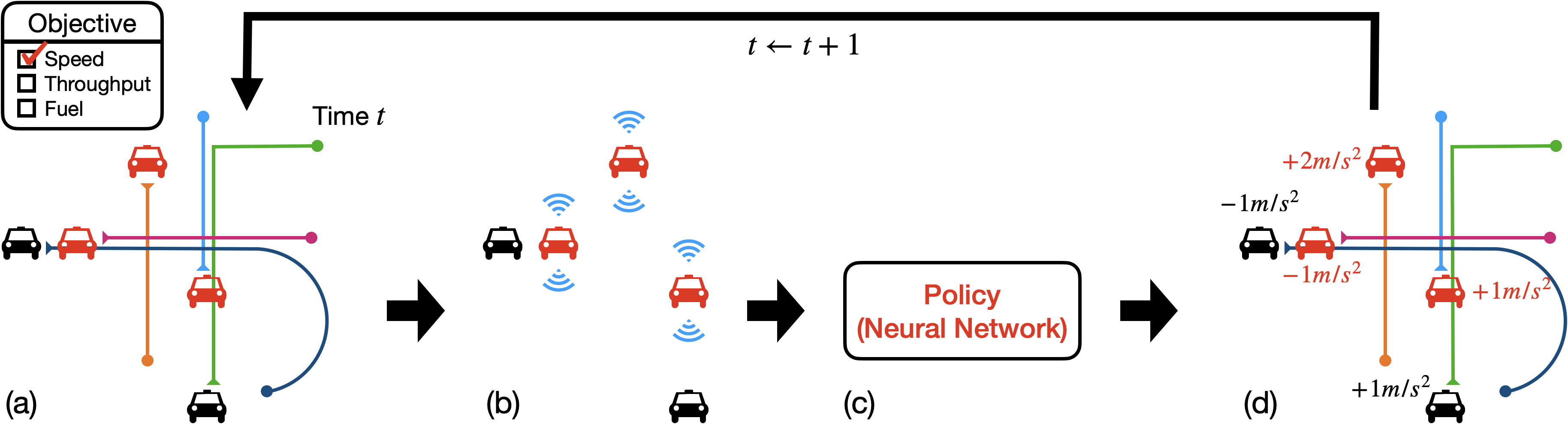}
\caption{\textbf{An exemplary vehicular system with an overview of our methodology.} a) At each time $t$, all vehicles have assigned routes towards assigned destinations; a fraction of vehicles are automated (red) while the rest are uncontrolled (black). We would like to control the AVs towards maximizing a desired objective. b) We define the set of neighboring vehicles that each AV may sense. c) Based on each automated vehicle's observed surrounding, a learned control policy dictates the action of each AV towards optimization of the objective. Uncontrolled vehicles follow some default policy. d) Position and speeds of all vehicles are updated, and the process repeats at time $t + 1$.}
\label{fig:traffic_systems}
\end{figure*}

\section{Introduction}
\IEEEPARstart{A}{} developing trend in mobility systems today is the full or partial adoption of automated control of mobile vehicles in traditionally human-operated roles \cite{morris2017guest}. This trend can be observed in systems ranging from real-world traffic systems \cite{wadud2016help} to warehouses employing mobile robots for storage, sorting, or delivery \cite{wurman2008coordinating}. Increasing autonomy in these systems \edit{increases} the potential to algorithmically control and coordinate automated vehicles (AVs) to increase efficiency, reduce congestion, or optimize other objectives like fuel usage throughout the system.

For the near future, while AV adoption remains fractional, automated control in real-world traffic systems would necessarily interact with human control, creating \textit{mixed autonomy} traffic. While any control may be underactuated in the mixed autonomy setting, such control may still induce desired behaviors, as demonstrated by several recent works on reducing congestion in simulated, mixed autonomy circular \cite{wu2021flow} or highway traffic systems \cite{vinitsky2018lagrangian}.

In many cases, mixed or full automation must solve an underlying mixed discrete and continuous optimization problem, which may be difficult to even formulate due to complex and stochastic dynamics, let alone solve practically. For many such complex systems, simulation decouples modeling of the system from further analysis and optimization efforts. For this reason, researchers and practitioners construct simulations of varying fidelity for many real-world systems.

In this study, we demonstrate the generality and ease of applicability of a unified model-free deep reinforcement learning (DRL)-based methodology for optimizing behaviors in diverse mixed autonomy traffic systems in simulation. In contrast to planning and search algorithms, model-free reinforcement learning is applicable to both continuous and discrete domains and only requires the ability to simulate forward trajectories from a set of initial states; moreover, learned DRL policies may execute more efficiently in real-time and at scale.

We acknowledge that a sizeable gap exists between simulation and reality, especially in real-world traffic systems with many stochastic actors with human-intent. While preliminary Sim2Real policy transfer has been done in miniature physical traffic systems \cite{jang2019simulation}, we do not expect Sim2Real policy transfer to be feasible for real-world traffic systems in the near future \cite{hofer2021sim2real}. Nevertheless, we attempt to derive insights and interpretable controllers from the learned policies. Toward this end, previous works have designed simulated automation and control strategies for industrial warehouse vehicles \cite{digani2015ensemble}, metro train regulation \cite{lin2011optimization}, airport surface management \cite{morris2016planning}, container loading \cite{stavrou2017optimizing}, and even pedestrian control \cite{zhu2020optimal}. Moreover, we argue that near-future Sim2Real extensions of our work \textit{are feasible} for fully automated robotic systems, which may require movement and coordination of automated vehicular robots with assigned routes \cite{wurman2008coordinating}. These settings are suitable due to 1) existence of higher fidelity simulators, 2) little or no need to simulate human intent, and 3) existence of collision avoidance mechanisms.

This work follows a series of our previous works applying DRL to mixed autonomy traffic \cite{wu2021flow,wu2017emergent,kreidieh2018dissipating,vinitsky2018benchmarks,vinitsky2020optimizing,yan2021reinforcement}. While each previous work often focuses on a single traffic system and applies significant amounts of system-specific handcrafting (indeed, DRL-based methods in various applications are known to be notoriously hard to tune to good performance), this work presents a simplified and unified DRL methodology for a superset of open and closed traffic systems, with a focus on ease of applicability. The code introduced in this work is a lightweight revision of the Flow Framework \cite{wu2021flow}, completely rewritten to offer researchers and practitioners more control in designing the traffic system while minimizing the amount of DRL design choices and hyperparameters. Additionally, we interpret the behaviors of DRL-controlled AVs, some of which resemble those designed by traffic engineering experts, and design simple controllers inspired by the learned DRL policies. While our previous works may offer deeper insights into the performance of DRL in particular traffic systems, this work emphasizes the ease of applicability of model-free DRL to general vehicular systems with mixed or full autonomy.

In summary, the contributions of our present work are:
\begin{enumerate}
    \item We present a unified variable-agent, multi-task DRL methodology and showcase the generality, effectiveness, and ease of usage for optimizing mixed autonomy traffic in simulated vehicular systems.
    \item To shed light \edit{on} the performant behaviors discovered automatically via DRL, we manually extract and benchmark simple controllers inspired by the behaviors.
    \item We characterize the robustness of each trained policy across a range of vehicle densities.
    \item We introduce a lightweight codebase with heavy emphasis on ease of usage and simplified design choices for researchers and practitioners.
\end{enumerate}
Code, models, and videos of results \edit{are available on \href{https://github.com/mit-wu-lab/automatic_vehicular_control}{Github}}.

The rest of this article is organized as such: Section~\ref{sec:related_work} details related work, Section~\ref{sec:preliminaries} introduces relevant DRL concepts, Section~\ref{sec:automated_vehicular_systems} defines relevant vehicular systems, Section~\ref{sec:rl_methodology} details the DRL methodology, Section~\ref{sec:setup} discusses experimental setup, and Section~\ref{sec:results} provides experimental results.

\section{Related Work}
\label{sec:related_work}
\textbf{Traffic control.} Due to the ubiquity and costs of congestion in traffic, much work has been devoted to traffic control for reducing congestion and increasing efficiency \cite{papageorgiou2003review}. In urban traffic networks, composed of many intersections, traffic signal control strategies have been widely studied and sometimes deployed for isolated or coordinated intersections, including fixed-time \cite{little1981maxband} or adaptive \cite{hunt1981scoot}. In freeway traffic networks \cite{treiber2013traffic}, ramp metering control methods like ALINEA \cite{papageorgiou1991alinea} are deployed to counter congestion due to reduction in road capacity, and cooperative adaptive cruise control (CACC) \cite{van2006impact} methods are proposed to mitigate congestion due to perturbations in traffic flow. While works in CACC and our study both concern vehicular control of traffic, we aim to automatically discover optimal traffic behavior rather than manually prescribing desired speeds for vehicles to follow.

\textbf{Isolated autonomy.} The US DARPA challenges in autonomous driving spurred much research in components necessary for the deployment of automated vehicles in the real-world \cite{buehler2009darpa}. These developments, along with developments in advanced driver assistance systems (ADAS) like adaptive cruise control (ACC) \cite{vahidi2003research}, focus on the safety, comfort, ``human-ness,'' and performance of an individually automated vehicle rather than the traffic system as a whole \cite{lefevre2015learning}.

\textbf{Mixed and full autonomy.} Unlike isolated autonomy, mixed and full autonomy are often studied as traffic control techniques aimed to optimize local or system-wide objectives. CACC methods \cite{van2006impact} are often studied under mixed and full autonomy settings with varying penetration rates of CACC vehicles. Mixed and full autonomy control of intersections have been studied by \cite{sharon2017protocol} and \cite{dresner2008multiagent,wu2012cooperative,miculescu2019polling}, respectively. However, the former analyzes the performance of a fixed first-come-first-serve protocol while the latter abstracts away intersection dynamics into a polling-system of two queues. As discussed in more detail in \cite{wu2021flow}, two prominent challenges in studying mixed autonomy in particular are the high uncertainty in system dynamics, due to modeling human behavior, and the lack of a known optimal behavior. As we show in this work and our previous works, DRL may be a suitable methodology which addresses both challenges. \edit{Throughout this work, we assume that all control decisions are supported by ideal vehicular communication, and we defer to \cite{wang2020thirty,ko2018wireless} for a discussions realistic, non-ideal communication in vehicular systems.}

\textbf{Model-free reinforcement learning.} Derived from optimal control and machine learning, model-free reinforcement learning (RL) is a methodology for optimizing sequential decision making \cite{bertsekas1996neuro,sutton2018reinforcement}. Model-free RL decouples optimal control from system modeling with Markov Decision Process as the interface. Addressing the challenges of mixed autonomy, model-free RL does not need to model the dynamics of the underlying system and does not require knowledge of an optimal behavior. Recently, model-free DRL, combining model-free RL with deep neural networks, has demonstrated improved performance for traffic signal control \cite{wei2019colight}, ramp metering \cite{belletti2017expert}, and multi-robot navigation \cite{liu2021visuomotor}. Other applications of model-free DRL to automation include optimal parameters for computer numerical control (CNC) machining \cite{xiao2019meta} and optimal scheduling in manufacturing \cite{park2019reinforcement,ou2020method}. However, as model-free DRL algorithms are primarily simulation-based, deployment in real-world settings suffers from several difficulties \cite{hofer2021sim2real}; in this article, we briefly acknowledge the gap between simulation and deployment.

\textbf{Model-free DRL for mixed autonomy traffic.} This work generalizes our previous works on applications of model-free DRL to mixed autonomy traffic systems based on the Flow framework \cite{wu2017framework,wu2021flow,wu2017emergent,vinitsky2018benchmarks,vinitsky2018lagrangian,kreidieh2018dissipating,vinitsky2020optimizing,yan2021reinforcement}. While each previous work demonstrates that DRL overcomes long-standing classical control challenges in traffic control, including complex dynamics models, long horizons, partial observations, and non-standard noise, these work often included artificial encouragement and handcrafting to guide the DRL policy in their specific traffic system.

For reward shaping, \cite{wu2017framework} and \cite{wu2021flow} penalize acceleration and deceleration to encourage convergence to a constant speed in ring-like traffic systems while \cite{wu2017emergent,kreidieh2018dissipating,vinitsky2018benchmarks} penalize deviation from tuned desired speed hyperparameters in figure-eight and highway ramp traffic systems. On the other hand, \cite{vinitsky2018lagrangian} and \cite{vinitsky2020optimizing} restricts control over AVs to selected segments of a highway bottleneck system to encourage ramp metering-like behavior. In both cases, reward shaping and selective control of vehicles not only require cumbersome tuning the researcher or practitioner, but are constrained by human intuition which may be suboptimal in more complex systems. Moreover, \cite{wu2017framework,wu2021flow,wu2017emergent,vinitsky2018benchmarks,vinitsky2018lagrangian,kreidieh2018dissipating,vinitsky2020optimizing} all use actor-critic algorithms ranging from TRPO \cite{schulman2015trust} to TD3 \cite{fujimoto2018addressing} which require extensive hyperparameter tuning for training neural networks for both the policy and a value function critic; in contrast, this work shows that the TRPO algorithm with only a policy neural network and without a value function network suffices for all considered traffic systems, eliminating much of the hyperparameter tuning.

Overall, this work shows that a unified DRL methodology achieves similar or better efficiency without resorting to system-specific hand-designing to ease optimization. We believe that the ability to easily discover performant behaviors in any system without hand-holding is key towards broader applicability of DRL in general vehicular systems.

% Bottleneck:
%     https://flow-project.github.io/papers/LagrangianBottlenecks.pdf
%     https://arxiv.org/pdf/2011.00120.pdf
% Open highway / ramp and closed loops
%     https://flow-project.github.io/papers/08569485.pdf
% Figure 8, merge, bottleneck, grid (traffic light)
%     http://proceedings.mlr.press/v87/vinitsky18a/vinitsky18a.pdf
% Ring, multilane ring, figure 8 (1 av and 14 av)
%     http://proceedings.mlr.press/v78/wu17a/wu17a.pdf
% Ring
%     https://ieeexplore.ieee.org/document/9489303

\section{Preliminaries}
\label{sec:preliminaries}
\subsection{Markov Decision Process (MDP)}
Markov Decision Process (MDP) is a framework for modeling sequential decisions. We model each decision process in this paper as a finite-horizon discounted MDP, defined by $\cM = (\cS, \cA, T, r, \rho_0, \gamma, H)$ consisting of state space $\cS$, action space $\cA$, stochastic transition function $T(s, a, s') = p(s'|s, a)$ for $s, s' \in \cS$ and $a \in \cA$, reward function $r(s, a, s') \in \bbR$, initial state distribution $\rho_0$, discount factor $\gamma \in [0, 1]$, and horizon $H \in \bbZ_+$. Given this MDP definition, reinforcement learning and optimal control aim towards the following objective
\begin{equation}
    \max_{a_0 \dots a_{H - 1} \in \cA} \bbE_{\substack{s_0\sim \rho_0,s_{t+1}\sim T(s_t, a_t, \cdot)}}\left[\sum_{t=0}^{H - 1} \gamma^t r(s_t, a_t, s_{t+1})\right]
\end{equation}
which maximizes the expected cumulative reward by selecting optimal actions $a_0 \dots a_{H - 1} \in \cA$.

While the vehicular control decision processes that we consider in this paper may instead be considered as infinite-horizon MDPs, which maximizes the expected cumulative reward for $H \to \infty$, practical methods often optimize over a large finite $H$ as a proxy for $H \to \infty$.

\subsection{Policy-based Model-free Deep Reinforcement Learning}
\label{sec:policy_based_model_free_rl}
Policy-based model-free DRL algorithms define a policy $\pi_\theta(a|s)$ which gives the probability of taking action $a\in \cA$ at state $s\in\cS$. The policy is parameterized by $\theta$ (\textit{e.g.} weights in a linear function or neural network), which is optimized so that the policy maximizes the expected cumulative reward
\begin{equation}
    \max_{\theta} \bbE_{\substack{s_0\sim \rho_0,a_t\sim \pi_\theta(\cdot|s_t)\\ s_{t+1}\sim T(s_t, a_t, \cdot)}}\left[\sum_{t=0}^{H - 1} \gamma^t r(s_t, a_t, s_{t+1})\right]
    \label{eq:pg_objective}
\end{equation}

The REINFORCE policy gradient algorithm \cite{williams1992simple} maximizes the expected cumulative reward of policy $\pi_\theta$ in Equation~\ref{eq:pg_objective} by sampling trajectory $(s_0, a_0\dots, s_{H-1}, a_{H-1}, s_H)$ and optimizing the following objective
\begin{equation}
    \arg\max_{\theta} \sum_{t=0}^{H - 1}\sum_{t'=t}^{H - 1} \gamma^{t' - t} r(s_{t'}, a_{t'}, s_{t' + 1})
    \label{eq:reinforce_objective}
\end{equation}
via gradient descent on $\theta$. At each update step
\begin{equation}
    \theta\leftarrow \theta + \alpha \sum_{t=0}^{H - 1} \nabla_\theta \left.\log \pi_\theta(\cdot|s_t)\right\vert_{a_t}\sum_{t'=t}^{H - 1} \gamma^{t' - t} r(s_{t'}, a_{t'}, s_{t' + 1})
    \label{eq:reinforce_update}
\end{equation}
where $\alpha$ is the learning rate.

Like REINFORCE, the Trust Region Policy Optimization (TRPO) algorithm \cite{schulman2015trust} also collects a trajectory and optimizes the objective in Equation~\ref{eq:reinforce_objective}. However, to encourage training stability, TRPO constrains the update step of $\theta$ so that the policy does not change too quickly:
\begin{equation}
\begin{split}
    \theta\leftarrow \arg\max_{\theta'} \sum_{t=0}^{H - 1} \frac{\pi_{\theta'}(a_t|s_t)}{\pi_\theta(a_t|s_t)}\sum_{t'=t}^{H - 1} \gamma^{t' - t} r(s_{t'}, a_{t'}, s_{t' + 1})&\\ \text{ subject to } \frac{1}{H} \sum_{t=0}^{H - 1} D_\text{KL}(\pi_\theta(\cdot|s)\| \pi_{\theta'}(\cdot|s)) \leq \delta_\text{kl}
    \label{eq:trpo_update}
\end{split}
\end{equation}
where $\delta_\text{kl}$ is the upper bound of the mean KL divergence between the updated policy $\pi_{\theta'}$ and the original policy $\pi_\theta$, preventing $\theta'$ from deviating too far from $\theta$. In practice, TRPO eliminates the need to tune the learning rate $\alpha$, which is a sensitive hyperparameter while $\delta_\text{kl}$ is a standard constant.

We note that unlike our previous works \cite{wu2021flow,vinitsky2020optimizing}, we solely learn a policy and do not additionally learn a value function. \edit{Many actor-critic and value-based algorithms learn a value function or Q-value function to fit the cumulative reward at a given state or a given state-action, respectively. While these algorithms demonstrate improved performance in certain applications, properly learning a value function requires selection of the optimizer (\textit{e.g.} ADAM~\cite{kingma2014adam}, RMSprop~\cite{hinton2012neural}), tuning of the learning rate, tuning of the smoothing hyperparameter $\lambda$ for generalized advantage estimation~\cite{schulman2016high}, tuning of the hyperparameter for value clipping, and other potentially difficult algorithmic choices. In multi-vehicle domains in particular, fitting the value function to the system objective likely requires the value function to operate over the entire state at a given time, which is difficult and inflexible to encode, while we specify in Sections~\ref{sec:partial_observability}~and~\ref{sec:multiagent_decomposition} that our policy-based methodology only requires local observability.}

\section{Automated Vehicular Systems}
\label{sec:automated_vehicular_systems}
We describe the general types of simulated vehicular systems compatible with our methodology for automatic vehicle control. Overall, we focus on microscopic simulations which consider the interactions of individual vehicles rather than aggregate behavior of traffic flow. We require the ability to repeatedly run simulations for a duration from a set of initial simulation states. Each simulation evolves the positions and velocities of the vehicles through time following defined physical rules. We assume that every vehicle in the system follows its own route, which is assigned by some fixed algorithm given the origin and destination; we do not consider decision-making for route assignment in this work. In \textit{closed} systems, vehicles circulate within the system endlessly, following assigned routes. In \textit{open} systems, vehicles enter the systems (\textit{inflow}) at their origins and exit the systems (\textit{outflow}) at their destinations. Within each system, a fraction or all of the vehicles are automated and can be controlled in some manner, while the rest of the vehicles follow modeled default behavior; each system must have one or more automated vehicles. We assume that a central objective exists and can be quantified for the system; for example, the objective could be a function of vehicle speeds, system throughput, fuel consumption, or safety in the system. Note that even for system objectives purely based on speeds or throughput, attempting to control each \textit{individual} vehicle towards the maximum speed possible could often be suboptimal due to negative, congestion-inducing effects on surrounding vehicles.

We briefly describe two such automated vehicular systems:
\subsubsection{Traffic Systems}
Mixed autonomy traffic systems where fractions of all vehicles are automated are studied in \cite{wu2017framework,wu2021flow,wu2017emergent,vinitsky2018benchmarks,vinitsky2018lagrangian,kreidieh2018dissipating,vinitsky2020optimizing,yan2021reinforcement}. The acceleration and lane change (if applicable) decisions of AVs may be controlled. Uncontrolled vehicles follow default behavior dictated by well-studied car-following models, such as the Intelligent Driver Model (IDM)~\cite{treiber2000congested}.

\subsubsection{Robotic Warehouse Systems}
A warehouse management system for hundreds of cooperatively controlled mobile robots with origins and destinations is studied in \cite{wurman2008coordinating}. In this system, routing and movement of robots comprise a challenging joint control task which is often decoupled into dynamic route assignment followed by movement planning. Dynamic route assignment is addressed by \cite{digani2015ensemble, liu2020prediction} and other methods, while our methodology for automated vehicle control applies to the movement planning problem. Acceleration of the robots may be controlled. Strategies resembling mixed autonomy may reduce the computational complexity: control may be restricted to the set of robots making critical decisions at a given time while non-critical robots can follow default behaviors (\textit{e.g.} move at the maximum speed possible).

Other examples of such systems include metro train regulation \cite{lin2011optimization} and airport surface management \cite{morris2016planning}.

In this article, we validate the methodology on traffic systems. In practice, each system may support a variety of vehicle densities. Therefore, we desire policies which generalize across multiple \textit{configurations} of vehicle density.

\section{Deep Reinforcement Learning Methodology}
\label{sec:rl_methodology}
We prescribe a unified DRL methodology for automatic control of vehicular systems. An important contribution of this work is to minimize the DRL-related design choices that a researcher or practitioner has to make. Here we describe the common components of the DRL methodology necessary for all vehicular systems.

\subsection{MDP Definition}
We naturally model a vehicular system in microscopic simulation as a MDP. At each time step $t$, the state $s_t$ is composed of the positions, velocities, and other metadata of all vehicles in the road network. The action $a_t$ is the tuple of per-AV actions of all AVs in the road network at step $t$. The reward function $r(s_t, a_t, s_{t+1})$ is specified so that the cumulative reward is the objective. A key distinction from our previous works is that the reward function does not need to be manually shaped by the researcher or practitioner to encourage behaviors attaining higher objective values. The stochastic transition function $T$ is not explicitly defined, but $s_{t+1} \sim T(s_t, a_t, \cdot)$ can be sampled from the microscopic simulator, which applies the actions for all vehicles over a simulation step duration $\delta$. In simulations which do not protect against vehicle collisions, we introduce a collision penalty $-\lambda_\text{collision} n_\text{collision}$ to the reward function $r(s_t, a_t, s_{t+1})$ where $n_\text{collision}$ is the number of collided vehicles in $s_{t+1}$ and $\lambda_\text{collision}$ is large enough to discourage the AVs from collision-inducing behaviors.

\subsection{Partial Observability}
\label{sec:partial_observability}
In practice, as the state $s$ could be large or difficult to reason about, DRL methods often approximate the policy with $\pi_\theta(\cdot|s) \approx \pi_\theta(\cdot|o)$ \cite{schulman2015trust}, where the \textit{observation} $o = z(s) \in \cO$ possesses only a subset of the information of the state $s$, $z$ is the observation function, and $\cO$ is the observation space. Together, $(\cM, \cO, z)$ actually defines a partially observable MDP (POMDP) \cite{kaelbling1998planning}. Partial observability is natural in real-world decision processes since obtaining a local observation may be more feasible than the full state. For example, an AV may more easily observe nearby vehicles than faraway vehicles, and information regarding faraway vehicles may not help the decision algorithm much anyways. Guided by these principles, we design the observation function for systems with a single AV to include only the entities most relevant to the AV's decision; systems with multiple AVs will be considered next. The observation function is one of the few design choices made by the researcher or practitioner for the methodology presented within this work.

\subsection{Multi-agent Policy Decomposition}
\label{sec:multiagent_decomposition}
In vehicular systems with multiple AVs, we apply multi-agent policy decomposition with each AV as an \textit{agent}. A MDP with multiple action dimensions could naturally be formulated as a decentralized partially observable MDP (Dec-POMDP) \cite{boutilier1996planning}. In this case, we refer to the action space of the original MDP as the joint action space, which factorizes into the product of $M$ agent action spaces in the Dec-POMDP framework. The policy $\pi_\theta(a|s)$ decomposes into per-agent policies $\pi_\theta(a^i|o^i)$ such that $\pi_\theta(a|s) = \prod_{i = 1}^M \pi_\theta(a^i|o^i)$, where $a^i \in \cA^i$, the action space of agent $i \in \{1, \cdots, M\}$, and $o^i = z(s, i) \in \cO^i$, the observation space for agent $i$. We have $o^1 \cup \dots \cup o^M \subseteq s$ and $\cA^1 \times \dots \times \cA^M = \cA$. $z$ is a defined observation function which maps state $s$ to observation $o^i$ for agent $i$. Without multi-agent policy decomposition, the combinatorial nature of $\cA$ poses an intractable problem to learning algorithms. Like in single-AV systems, $z$ must be designed by the researcher or practitioner.

\subsection{Per-AV Action Space}
The longitudinal per-AV action space can often be naturally formulated as a continuous acceleration space $\cA_\text{longitudinal}^i = [-c_\text{decel}, c_\text{accel}]$ for each AV $i$. However, in systems where multiple AVs interact, we prescribe a discrete bang-off-bang acceleration space $\cA_\text{longitudinal}^i = \{-c_\text{decel}, 0, c_\text{accel}\}$, which we find to empirically improve coordination between multiple AVs despite forgoing fine-grained control. For systems which require AVs to make lateral (lane change) decisions, the lateral action space is the set of lane indices $\cA_\text{lateral}^i = \{1, \dots, L\}$ to travel in, where $L$ is the number of lanes. Therefore $\cA^i = \cA_\text{longitudinal}^i \times \cA_\text{lateral}^i$ for systems permitting AV lane change and $\cA^i = \cA_\text{longitudinal}^i$ otherwise.

\subsection{Per-AV Policy Architecture}
We define the per-AV policy $\pi_\theta(a^i|o^i)$ as a neural network with three fully-connected layers with a hidden size of 64. If there are multiple AVs, we share the policy parameter $\theta$ across all vehicles in the traffic network to share experiences between AVs \cite{gupta2017cooperative}. For systems requiring joint action for each AV (\textit{i.e.} $\cA^i = \cA_\text{longitudinal}^i \times \cA_\text{lateral}^i$), the policy is a neural network with multiple heads, one for each factor of the joint action.

\subsection{Reward Centering and Normalization}
To reduce variance of policy gradients \cite{engstrom2019implementation}, we apply reward centering and normalization to the original reward before calculating the objective for policy gradient
\begin{equation}
r_\text{norm}(s_t, a_t, s_{t+1}) = \frac{r(s_t, a_t, s_{t+1}) - \hat \mu_{r}}{\hat \sigma_{R}}
\end{equation}
where $\hat \mu_{r}$ is the running mean of $r$ and $\hat \sigma_{\hat R}$ is the running standard deviation of the running cumulative reward, which is updated according to $\hat R \leftarrow \gamma \hat R + r(s_t, a_t, s_{t+1})$.

\subsection{Multi-task Learning over Configurations}
As we consider multiple configurations with varying vehicle densities for each vehicular system, training a separate policy for each configuration would be cumbersome and inefficient. Thus, we discretize the density configuration space into equally-spaced density configurations and learn a single multi-task policy over this configuration set. During training, we initialize separate environments for each configuration in the configuration set. At each training step, our policy gradient algorithm receives trajectories from all environments and batches the gradient update due to these trajectories. Multi-task learning allows a single trained policy to generalize across a range of configurations, avoiding the costs of training a separate policy for each configuration.

\subsection{Derived Policies}
We extract the behaviors discovered by DRL policies by hand-designing simple rule-based policies with one or two optimized parameters. We denote these policies as the \textit{Derived} policies because they are grounded in the DRL policies' behaviors. The purpose of constructing Derived policies is two-fold: 1) the Derived policies offers a comparison between the DRL policy and a gold-standard policy which shares the similar behavior 2) the Derived policies are easily interpretable and may be analyzed further for practical deployment. \edit{The steps we take to construct a Derived policy are as follows:\begin{enumerate}
    \item Given a traffic system, train a performant DRL policy over a desired range of density configurations.
    \item For each density configuration, examine the behavior of the DRL policy through videos and time-space diagrams, noting common behaviors shared across multiple density configurations.
    \item Formulate these behaviors into an interpretable rule-based parameterized policy across density configurations.
    \item For each density configuration independently, obtain the optimal parameters of derived policy with exhaustive grid search, which is feasible and straightforward for low-dimensional parameter spaces.
\end{enumerate}}
To ease the hand-design process, we permit Derived policies to use information from any part of the state, contrasting with DRL policies which must arrive at decisions based on observed information only and must generalize well across all density configurations. When possible, we identify optimal parameter values that may be shared across ranges of densities configurations.
% Each Derived policy is optimized conditioned on a particular density configuration, and Derived policies may use information from any part of the state; independently for each density configuration, we perform an exhaustive search over the parameter space to optimize the Derived policy. These relaxations ease the hand-design process but contrast with the assumptions for a DRL policy, which must arrive at decisions based on observed information only and must generalize well across all density configurations.

\subsection{\edit{Complexity Analysis}}
\edit{We analyze the computational complexity of our methodology for vehicular systems with $M$ AVs. Our methodology decomposes the action space $\cA$ into the product of agent action spaces $\cA^i$ and restrict state space $\cS$ into corresponding agent observation spaces $\cO^i$. For our analysis, we make the assumption that the design of observation function $z$ preserves sufficient observability for decision making and that the decisions of other AVs do not induce non-stationarity. The former assumption may be realistic if observation functions are designed to exclude distant and non-causal facets of the state which may have little impact on the current decision. The latter assumption may be realistic if the policy changes slowly across updates, if individual acceleration actions have small effects on the state, and if the AV penetration rate is low. These assumptions factors the joint control problem to $M$ independent control problems, each with state space $\cO^i$ and action space $\cA^i$.}

\edit{The TRPO algorithm leveraged by our methodology is a variation of the Natural Policy Gradient (NPG) algorithm~\cite{kakade2001natural,agarwal2021theory}, so we proceed to invoke the computational complexity of NPG at training time. As proved by \cite{agarwal2021theory}, NPG obtains an $\epsilon$-optimal policy in tabular RL settings with no more than $\frac{2}{(1 - \gamma)^2\epsilon}$ gradient update steps, and each update step in the tabular case takes $O(|\cO^i||\cA^i|)$ time. In our work, we leverage functional approximation with a neural network, enabling tractable parameter updates though losing optimality guarantees of the tabular case. Updating the parameters of the neural network with a forward and backward pass of the neural network takes $f(|o|, |a|)$ time in general, where $o \in \cO^i$ and $a \in \cA^i$ are individual observation and action vectors, respectively. In practice, rather than defining an $\epsilon$ and taking $\frac{2}{(1 - \gamma)^2\epsilon}$ gradient update steps, we take a total of $G$ gradient update steps sufficient for the training performance to converge while batching over $B$ sampled trajectories for each gradient update, horizon $H$ simulation steps per trajectory, and $M$ agents at every simulation step, for a total training time complexity of $O(GBHMf(|o|, |a|))$. When executing an already trained model to generate a trajectories for $H$ simulation steps, the execution time complexity is $O(HMf(|o|, |a|))$.}

\edit{The $f(|o|, |a|) = O(|o| + |a|)$ computations introduced by small fully-connected networks such as ours may be small in practice compared to other factors such as inter-process communication overheads and the simulation time of the system itself, which is $O(GBHN)$ at training time and $O(HN)$ at execution time, where $N\geq M$ is the total number of vehicles in the system. However, if more computationally intensive convolutional, recurrent, or attention-based neural networks are necessary, the neural network computation time may increase significantly, requiring either more parallelism in the forms of GPUs or more efficient feature engineering and architecture design to reduce computational cost.}

\edit{Mixed autonomy ($M < N$) directly reduces computational complexity compared to full autonomy ($M = N$) by reducing $M$, while also indirectly reducing the complexity by lowering non-stationarity present in the system, permitting the usage of smaller $G$, $B$, and $H$ factors; in other words, in full autonomy systems the outcome of an AV’s decision is more likely to be affected by decisions made by other AVs, where all nearby vehicles are AVs. Nevertheless, mixed autonomy control may sacrifice some degree of optimality compared to full autonomy, due to underactuation.}

\begin{figure*}[!hbt]
\centering
\includegraphics[width=\textwidth]{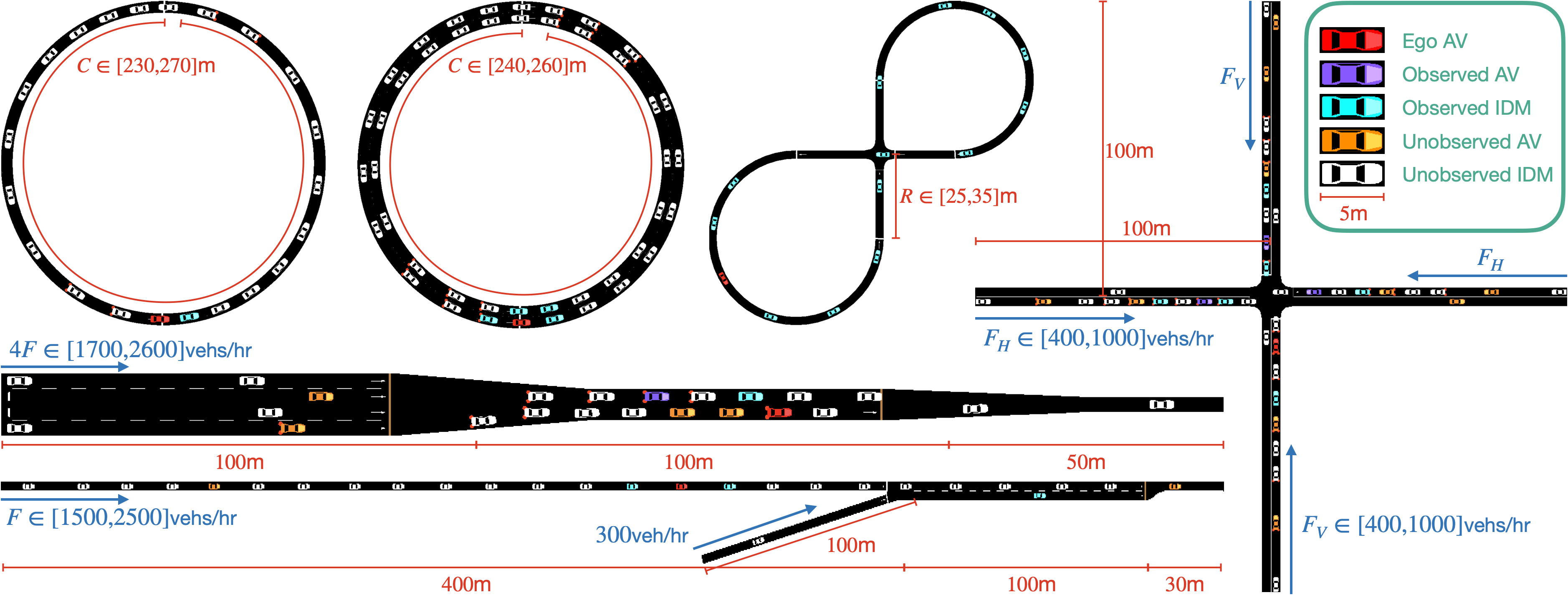}
\caption{\textbf{Experimental Traffic Systems.} In clockwise order from the top left: Single Ring, Double Ring, Figure Eight, Intersection, Highway Ramp, and Highway Bottleneck. Each traffic system is independently drawn to scale. Single Ring, Double Ring, and Figure Eight are closed systems with 22, 44, and 14 vehicles respectively. Intersection, Highway Ramp, and Highway Bottleneck are open systems with variable numbers of vehicles. Within each system, we designate one AV as the ego AV (red) without loss of generality and color other vehicles according to their type and whether they are observed by the ego AV. Each AV typically observes the speed, relative position, and type (AV or uncontrolled) of itself and its observed vehicles.}
\label{fig:traffic_systems}
\end{figure*}

\section{Experimental Setup}
\label{sec:setup}
We describe the general simulation, training, and evaluation setups of our experiments. We provide reference ranges here and reserve full details for the code.

\subsection{Vehicular Systems}
We construct six diverse mixed autonomy traffic systems in the SUMO microscopic simulator \cite{lopez2018microscopic} to demonstrate the generality of our unified methodology. Three systems are open and three systems are closed. While we do not incorporate any traffic control element, such as traffic light or ramp meter, future work may easily incorporate these traffic control elements as needed in conjunction with mixed autonomy control.

Common to all systems, all vehicles are 5m in length and uncontrolled vehicles follow the IDM with a Gaussian acceleration noise of 0.2m/s$^2$. A randomized initialization is obtained by simulating $H_0$ warmup steps starting from an arbitrary set of vehicle positions; the next $H$ steps are measured for performance. We use full SUMO safety checks, which prevent vehicles from entering most collisions situations. For each system, we consider multiple configurations of traffic densities.

In closed systems, we define the objective to be the total cumulative distance traveled by all vehicles, which is proportional to the average speed over all vehicles over all timesteps. We use a simulation step size of $\delta = 0.1$s for all closed systems and terminate the simulation immediately if an occasional collision occurs despite the safety check. The density configuration is varied by scaling the traffic network geometries while holding the number of vehicles constant.

In open systems, we designate the objective to be the throughput (outflow per hour) of the system. We use a simulation step size of $\delta = 0.5$s and do not terminate the simulation if two vehicles collide: only the collided vehicles are removed from \edit{the} simulation and do not count towards the outflow. Each density configuration corresponds to a target inflow rate (vehicles per hour), which controls the number of vehicles in the system. If a vehicle is not able to inflow due to congestion in an inflow lane, the vehicle is dropped from simulation. Unlike in closed systems, the number of vehicles in open systems is not constant and depends on the inflow rate.

As we already examine each traffic system under multiple traffic density configurations, we do not perform additional ablation on the effect of the AV penetration rate and instead choose to fix a penetration rate for each system. We refer readers to several previous works for ablations on the effect of penetration rate \cite{kreidieh2018dissipating,vinitsky2020optimizing}. While all of our systems contain under 50 vehicles at any given time, our methodology naturally extends to much larger systems if two conditions hold: 1) the objective (\textit{e.g.} throughput) can be decomposed into local objectives (\textit{e.g.} local throughput), and 2) the total number of decisions, which can be reduced via pruning, is not exorbitantly high.

We name and describe each traffic system, along with our constructed observation function. To encourage AVs to develop generalizable behaviors based on local information, we do not allow AVs to observe the underlying traffic density configuration parameter. All traffic systems and corresponding observation spaces are visualized in Figure~\ref{fig:traffic_systems}.

\subsubsection{Single Ring (Closed)}
\label{sec:setup_ring}
The Single Ring system consists of 22 vehicles in a single-lane ring network with circumference $C \in [230, 270]$~m; each $C$ corresponds to a density configuration. We designate one vehicle as AV while leaving the 21 other vehicles uncontrolled. The AV's observation function $z$ consists of the AV's speed and the offset and speed of the leading vehicle. We consider two differing objectives:

\paragraph{Global}
The objective is the cumulative distance traveled by all vehicles in the simulation. The reward function $r(s, a, s')$ is therefore the average speed of all vehicles in $s'$.

\paragraph{Greedy}
The objective is the cumulative distance traveled by the AV. The reward function $r(s, a, s')$ is therefore the AV's speed in $s'$.

\subsubsection{Double Ring (Closed)}
\label{sec:setup_double_ring}
The Double Ring system consists of 44 vehicles in a two-lanes ring network with circumference $C \in [240, 260]$~m. The SUMO simulator does not account for the exact geometry of the road and instead simulates the inner lane and outer lane to be the same length. We designate one vehicle in the outer lane as the AV, leaving the 43 other vehicles uncontrolled. In addition to controlling its own acceleration, the AV is allowed to change lane; no other vehicle is allowed to change lane. The observation function $z$ includes the speed and lane index of the AV and the speeds and offsets of the leading and following vehicles in both lanes. Like the Single Ring, we consider two cases corresponding to the $\{\text{Global}, \text{Greedy}\}$ reward functions.

\subsubsection{Figure Eight (Closed)}
\label{sec:setup_figure_eight}
The Figure Eight system consists of 14 vehicles in a closed single-lane two-way intersection network. Each direction (westbound or northbound) of the two-way intersection consists of length $R \in [25, 35]$~m straightaways before and after the intersection; each $R$ corresponds to a density configuration. The two directions are connected by 270$^\circ$ circular arcs. We designate one vehicle as AV, leaving others uncontrolled. The AV's observation function $z$ consists of the distance from the intersection and speed for every vehicle, reflecting the symmetry of the two loops. The reward function $r(s, a, s')$ is the average speed of all vehicles in $s'$.

\subsubsection{Highway Bottleneck (Open)}
\label{sec:setup_highway_merge}
The Highway Bottleneck system simulates a straight highway with four 100m-long inflow lanes which merge into two 100m-long lanes then merge into a single 50m-long lane, from which vehicles outflow. All four inflow lanes share a per-lane target inflow rate of $F$; so the total target inflow rate is $4F \in [1700, 2600]$~vehs/hr. At the first merge (four lanes to two lanes), the top two lanes merge together and the bottom two lanes merge together. No vehicle may change lane. We designate $20\%$ of the vehicles as AVs. Let the \textit{merge lane} be the lane which merges with the AV's lane. The observation function for each AV is the speed and the distance to the next merge of the AV, the offset and speed of closest following AV on the merge lane, and the offset and speed of closest following uncontrolled vehicle on the merge lane.

\subsubsection{Highway Ramp (Open)}
\label{sec:setup_highway_ramp}
The Highway Ramp system simulates a straight single-lane highway with an on-ramp. The single-lane highway proceeds for 400m when it meets a 100m on-ramp to form 100m of a two-lane merging region. The two-lanes merge into a single lane at the end of the 100m merging region, and the single-lane highway continues for another 30m. The highway sees a target inflow rate $F \in [1500, 2500]$~vehs/hr while the ramp sees \edit{a} target inflow rate of 300~vehs/hr. No vehicle may change lane. We designate $10\%$ of the highway vehicles as AVs, leaving the rest uncontrolled, including all ramp vehicles. The observation function for each AV is the speed of the AV, the offsets and speeds of the leading and following vehicles on the highway, and the offset and speed of the following vehicle on the ramp.

\subsubsection{Intersection (Open)}
\label{sec:setup_intersection}
The Intersection system simulates a single-lane intersection with inflows and outflows in each cardinal direction. The intersection only permits straight traffic and does not permit turns. Along each direction, the intersection is situated between two 100m long road segments. We consider configurations of pairs of horizontal and vertical target inflow rates $F_H, F_V \in [400, 1000]$~vehs/hr; configurations with $F_H + F_V < 1400$~vehs/hr are excluded due to trivially low inflow. We designate $33\%$ of the vehicles as AV. The observation function for each AV includes the position and speed of the heads and tails of the closest \textit{chains} to the intersection, where we define each \textit{chain} to be an AV and any uncontrolled vehicles that it immediately leads. The rationale behind this design is that each AV may provide control to all tailing uncontrolled vehicles.

\subsection{Baseline Policies}
For each system, we define the Baseline policy to follow the SUMO IDM behavior for all AVs. As collisions may frequent occur in the Figure Eight and Intersection systems under the Baseline policy, the vertical directions are given priority over the horizontal directions, which must slow to a near-stop before proceeding.

\edit{For the Single Ring, Highway Bottleneck, and Intersection systems, we adjust DRL algorithms from prior works~\cite{wu2021flow,vinitsky2018lagrangian,yan2021reinforcement} to train policies within our respective traffic systems, which are similar but may differ somewhat in construction from the those from the prior works. For these algorithms, we use the exact same training and evaluation setup as described below for our own methodology when applicable to ensure fairness of comparison.}

\subsection{Training}
For each system, we train a policy for up to $G = 200$ gradient update steps with the TRPO algorithm. We perform each gradient step with the batched data from $40 \leq B \leq 45$ collected trajectories, divided among equally-spaced configurations. For each trajectory, we use $H_0 \leq \frac{100}{\delta}$ warmup steps and horizon $H = \frac{1000}{\delta}$; warmup steps provide randomness in the MDP initialization. Unlike typical model-free DRL setups which may sweep over many DRL algorithms each with many hyperparameters involved in training the policy or value function, the only tuned hyperparameter in this article is the discount factor $\gamma \in [0.9, 0.9999]$, where $1 - \gamma$ is searched in log-space. Training each policy takes less than 3 hours on an Intel Xeon Platinum CPU machine with 48 cores. Though training is stochastic, we do not observe significant variations in learned behavior and performance between runs. For systems much larger than ones considered in this work, TRPO may result in slow training and high memory consumption and may be replaced with REINFORCE.

\subsection{Evaluation}
For each system, we select the checkpoint with the best average objective value on the batched training trajectories to evaluate. To evaluate the checkpoint on each configuration of each system, we sample 10 trajectories with different initial seeds. To allow traffic dynamics to achieve steady state, we use longer $H_0 \leq \frac{500}{\delta}$ warmup steps, sufficient to allow congestion to fully build up under the Baseline policy. We then run the policy for $H_1 \leq \frac{1500}{\delta}$ steps to allow traffic dynamics to achieve steady state under the evaluated policy, before measuring the objective value (speed or outflow) on a last $H \leq \frac{1000}{\delta}$ steps. The choice of $H_0$, $H_1$, and $H$ are not significant as long as $H_0$ and $H_1$ are each long enough for traffic dynamics to achieve steady state.

\section{Experimental Results}
\label{sec:results}
We present both numerical performance comparisons with the Baseline policy as well as a behavioral dissection, visualized via time-space diagrams, for representative configurations of each traffic system studied. We measure numerical performances after sufficient duration has passed for vehicle dynamics to achieve steady state. In all systems, we demonstrate that DRL discovers interesting and sometimes surprising behaviors which significantly outperform the Baseline. To shed light on the performant behaviors discovered automatically via DRL, we extract DRL behavior into rule-based Derived policies and offer numerical performance comparisons. For all speed and outflow results, we compute the means and standard deviations across 10 trajectories with different seeds; the standard deviation may sometimes be small for the Baseline and Derived policies. \edit{We reserve experiments demonstrating robustness of Derived policies to different car following model parameters for Appendix~\ref{sec:appendix}.}

\subsection{Single Ring}
\begin{figure}[!t]
\centering
\includegraphics[width=3.5in]{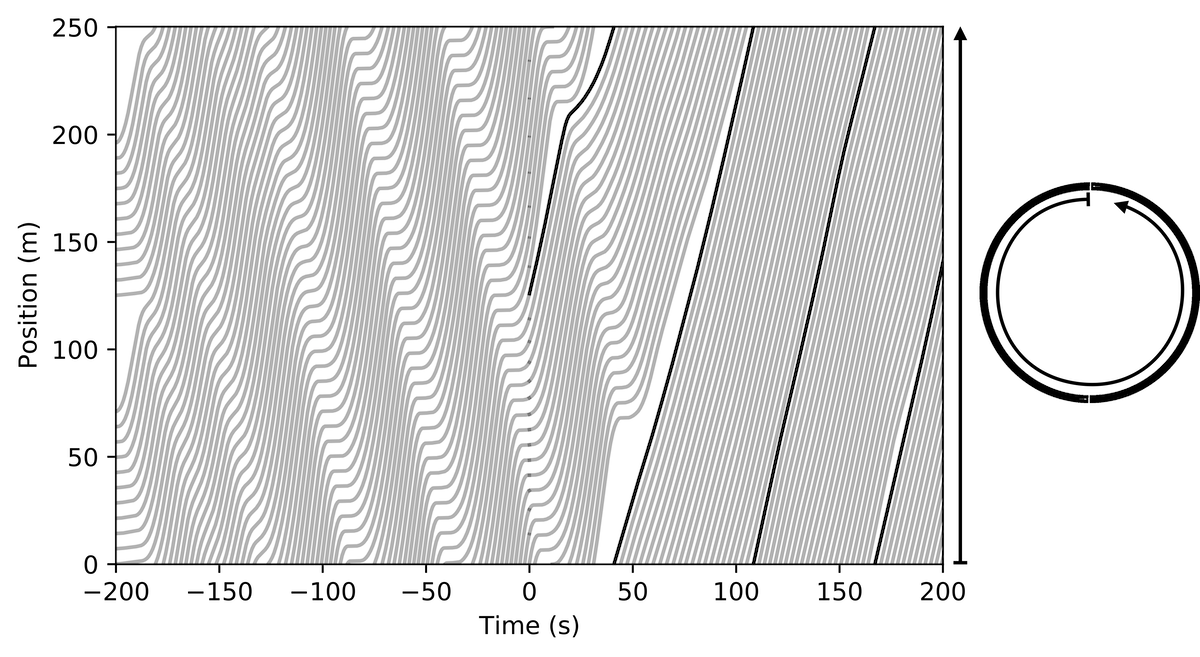}
\caption{\textbf{Single Ring $C = 250$~{\normalfont m} Time-space Diagram.} We plot the trajectories of vehicles under the Baseline policy (before time 0s) and the learned Global policy (on and after time 0s). Bold indicates the AV controlled by the DRL policy. Arrows indicate progression of vehicles. The DRL policy controls the AV to eliminate the backward propagating waves formed under the Baseline policy.}
\label{fig:ring_timespace}
\end{figure}

Due to the linear string instability of IDM \cite{herman1959traffic}, the Baseline policy quickly results in a stop-and-go waves which propagate in the opposite direction of traffic \cite{stern2018dissipation} under all density configurations. Under both the Greedy and Global policies, the AV learns to mitigate stop-and-go waves in every configuration by converging to a constant speed. We illustrate the Baseline and Global behaviors in Figure~\ref{fig:ring_timespace}.

\begin{figure}[!t]
\centering
\includegraphics[width=3.5in]{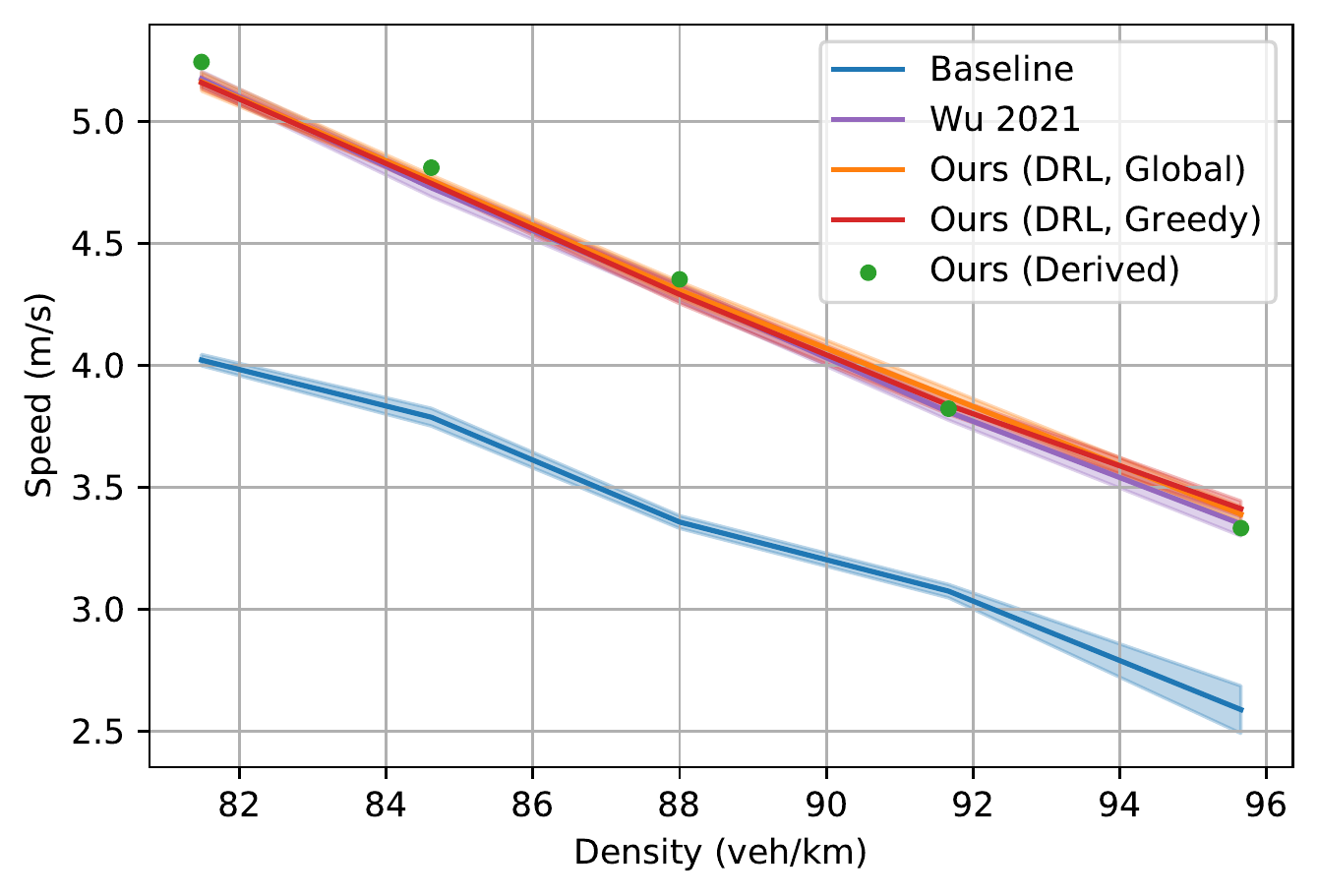}
\caption{\textbf{Single Ring Average Speed.} We compare the average speed over all 22 vehicles over horizon $H$ under the Baseline, DRL, and Derived policies\edit{, with the shaped deep reinforcement learning policy Wu 2021~\cite{wu2021flow} as additional comparison}. Our \edit{DRL} and Derived policies see significantly better average speeds than those of the Baseline policy. We display the Derived performance as points instead of a single line because the optimal speed parameter $v_\text{target}$ are not shared for any of the density configurations. \edit{We show that our unified methodology produces similar performances to Wu 2021 without hand-designed acceleration penalties which encourage convergence to a constant speed.}}
\label{fig:ring_speed}
\end{figure}

Mimicking this behavior, we design a Derived policy with a single, optimized target speed $v_\text{target}$ per circumference configuration. Figure~\ref{fig:ring_speed} compares the average speeds among the DRL, Derived, and Baseline policies. The DRL policies nearly matches the Derived policies despite seeing local observations only, without knowledge of the true circumference configuration. Our results here are similar to \edit{Wu 2021~\cite{wu2021flow} (Figure~\ref{fig:ring_speed})} with one important difference: the prior work utilizes an additional acceleration penalty to encourage convergent behavior in speed while we show that a simple speed-based objective alone is sufficient for DRL to discover convergent behavior. In addition, \cite{wu2021flow} only considers a global objective, while we consider both global and greedy objectives.
\begin{algorithm}
\caption{Single Ring Derived Policy}\label{alg:ring_derived}
\begin{algorithmic}[0]
\Procedure{Derived}{$s$} \Comment{State $s$}
\State $C \leftarrow$ get circumference from $s$
\State $v_\text{target} \leftarrow$ tuned target speed parameter for $C$
\State $v \leftarrow$ get speed of the AV from $s$
\State \Return Equalize($v_\text{target}, v$)
\EndProcedure
\\
\Procedure{Equalize}{$v_\text{target}, v_\text{current}$}
\If{$v_\text{current} < v_\text{target}$} \Return $0.75 c_\text{accel}$
\ElsIf{$v_\text{current} > v_\text{target}$} \Return $-0.75 c_\text{decel}$
\Else\ \Return 0
\EndIf
\EndProcedure
\end{algorithmic}
\end{algorithm}

\subsection{Double Ring}
\begin{figure}[!t]
\centering
\subfigure[Global Policy]{\includegraphics[width=3.5in]{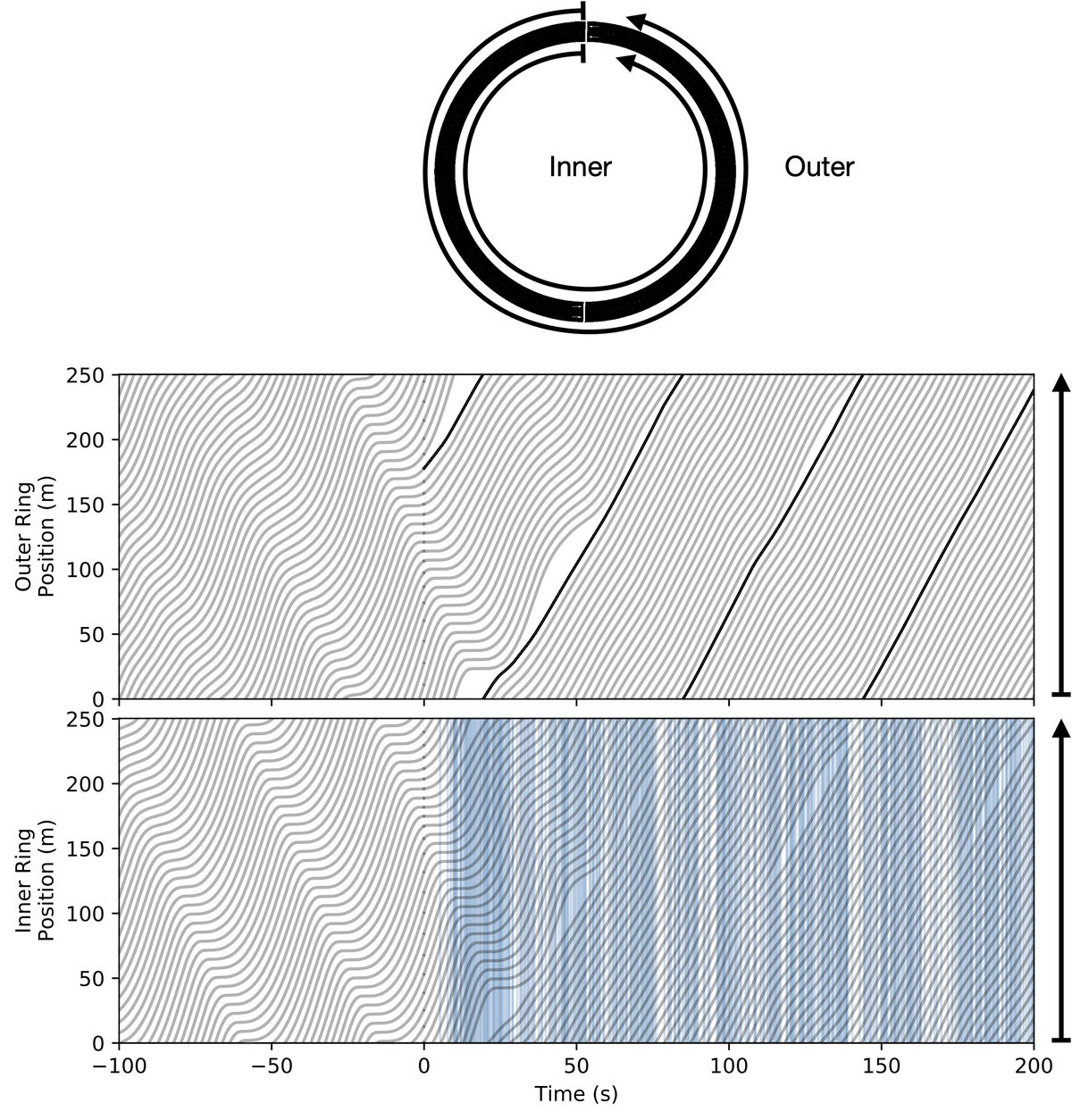}}
\subfigure[Greedy Policy]{\includegraphics[width=3.5in]{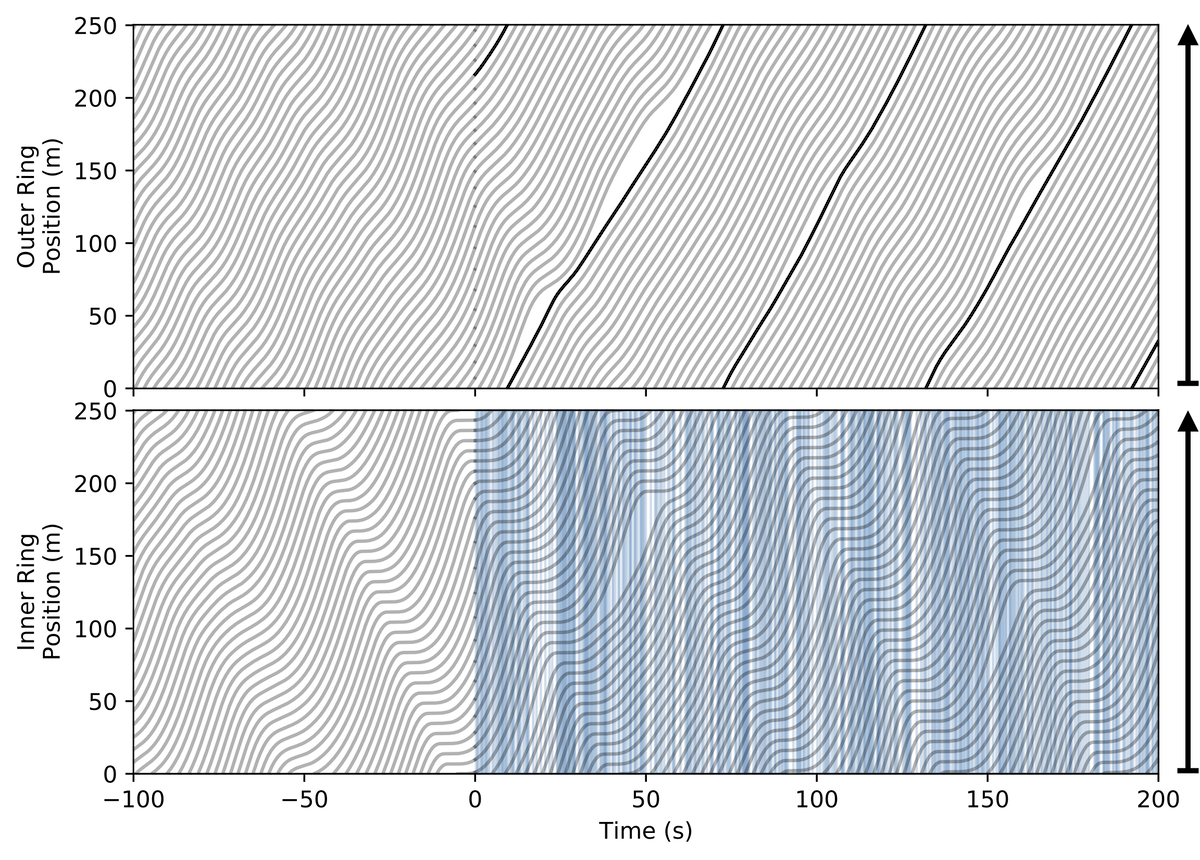}}
\caption{\textbf{Double Ring $C = 250$~{\normalfont m} Time-space Diagrams.} We plot the trajectories of vehicles under the Baseline policy (before time 0s) and the learned Global or Greedy policies (on and after time 0s). Bold indicates the AV controlled by the DRL policy. Arrows indicate progressions of vehicles in the outer and inner lanes. In both Global and Greedy, the DRL-controlled AV eliminates the backward propagating waves that form under the Baseline policy within its own lane. Turn signal flickering (blue vertical strips) by the Global policy strategically mitigates the waves that form in the \textit{other} lane, while that of the Greedy policy does not.}
\label{fig:double_ring_timespace}
\end{figure}

\begin{figure}[!t]
\centering
\includegraphics[width=3.5in]{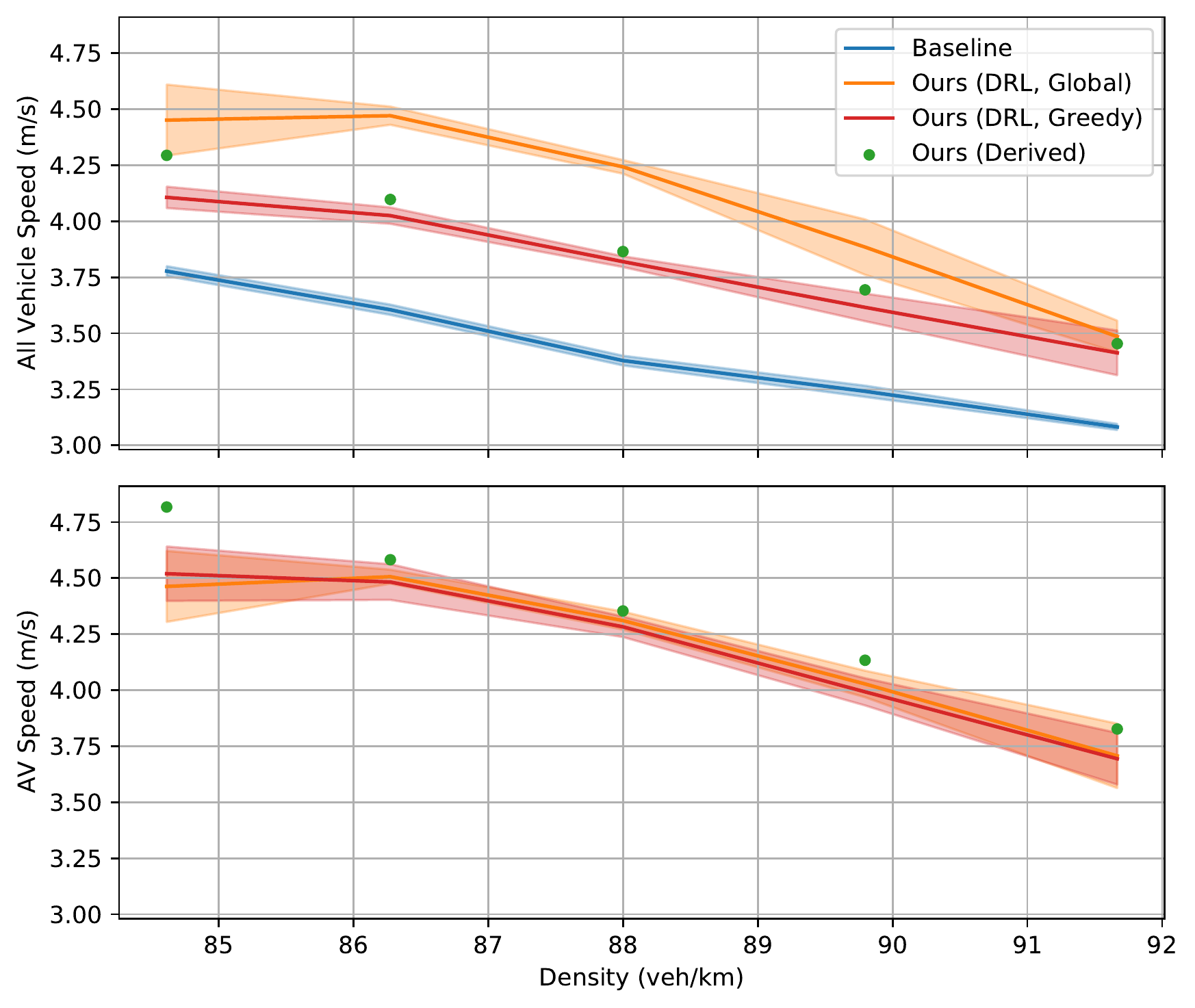}
\caption{\textbf{Double Ring Average and AV Speed.} Over horizon $H$, we compare the average speed over all 44 vehicles (top) and AV speed (bottom) under the Baseline, DRL, and Derived policies. The Global policy sees the best average speed in almost all cases, due to mitigation of stop-and-go waves in both lanes. The Derived and Greedy policies may see better AV speed than Global due to better mitigation of waves within the AV's own lane.}
\label{fig:double_ring_speed}
\end{figure}

Under the Baseline policy, each lane in the Double Ring exhibits identical behavior to the Single Ring. However, equipping the AV with the ability to change lane results in differing behaviors when maximizing the Greedy or Global objective with DRL. The Greedy policy learns to stay and converge to a constant speed within its own (outer) lane while disregarding the vehicle movement in the inner lane completely. On the other hand, the Global policy learns to mitigate the stop-and-go waves within both lanes \textit{simultaneously} by converging to a constant speed within its own lane while \textit{flashing the turn signal to regulate the speed of the inner lane without physically changing lane}. The behaviors are shown in Figure~\ref{fig:double_ring_timespace} and compared numerically in Figure~\ref{fig:double_ring_speed}. We note that the AV under the Greedy policy also frequently flickers its turn signal, as seen in Figure~\ref{fig:double_ring_timespace}; further investigation is required to differentiate the signal patterns of the two policies, which leads to significant differences in performance outcomes. Though this particular Global behavior exploits a flaw in the SUMO simulation, we note that a naturalistic human driver may also slow down if a leading vehicle in another lane attempts to change lane into the space ahead. We construct the Derived policy in an identical manner to the Single Ring, without lane change. However, our Derived policy lacks the strategic turn signal behavior of Global.

\subsection{Figure Eight}
\begin{figure}[ht]
    \centering
    \includegraphics[width=3.5in]{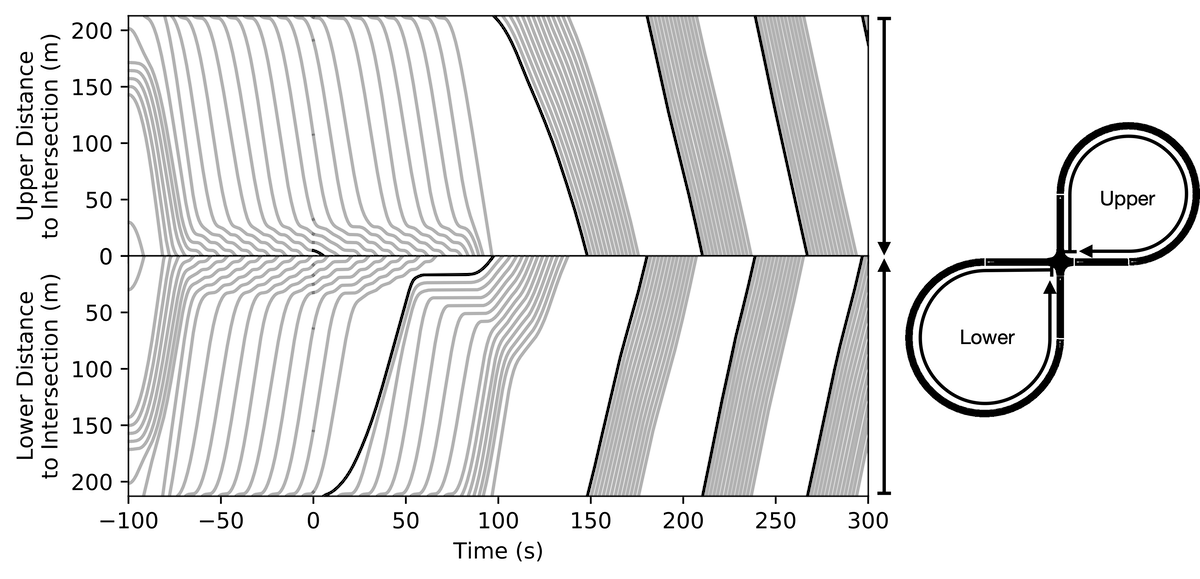}
    \caption{\textbf{Figure Eight $R = 30$~{\normalfont m} Time-space Diagram.} We plot the trajectories of vehicles under the Baseline policy (before time 0s) and the learned DRL policy (on and after time 0s). Bold indicates the AV controlled by the DRL policy. Arrows indicate progressions of vehicles approaching the intersection from the upper and lower loop. The AV guides a snaking behavior that eliminates alternation of single vehicles at the intersection.}
    \label{fig:figure_eight_timespace}
\end{figure}

\begin{figure}[!t]
\centering
\includegraphics[width=3.5in]{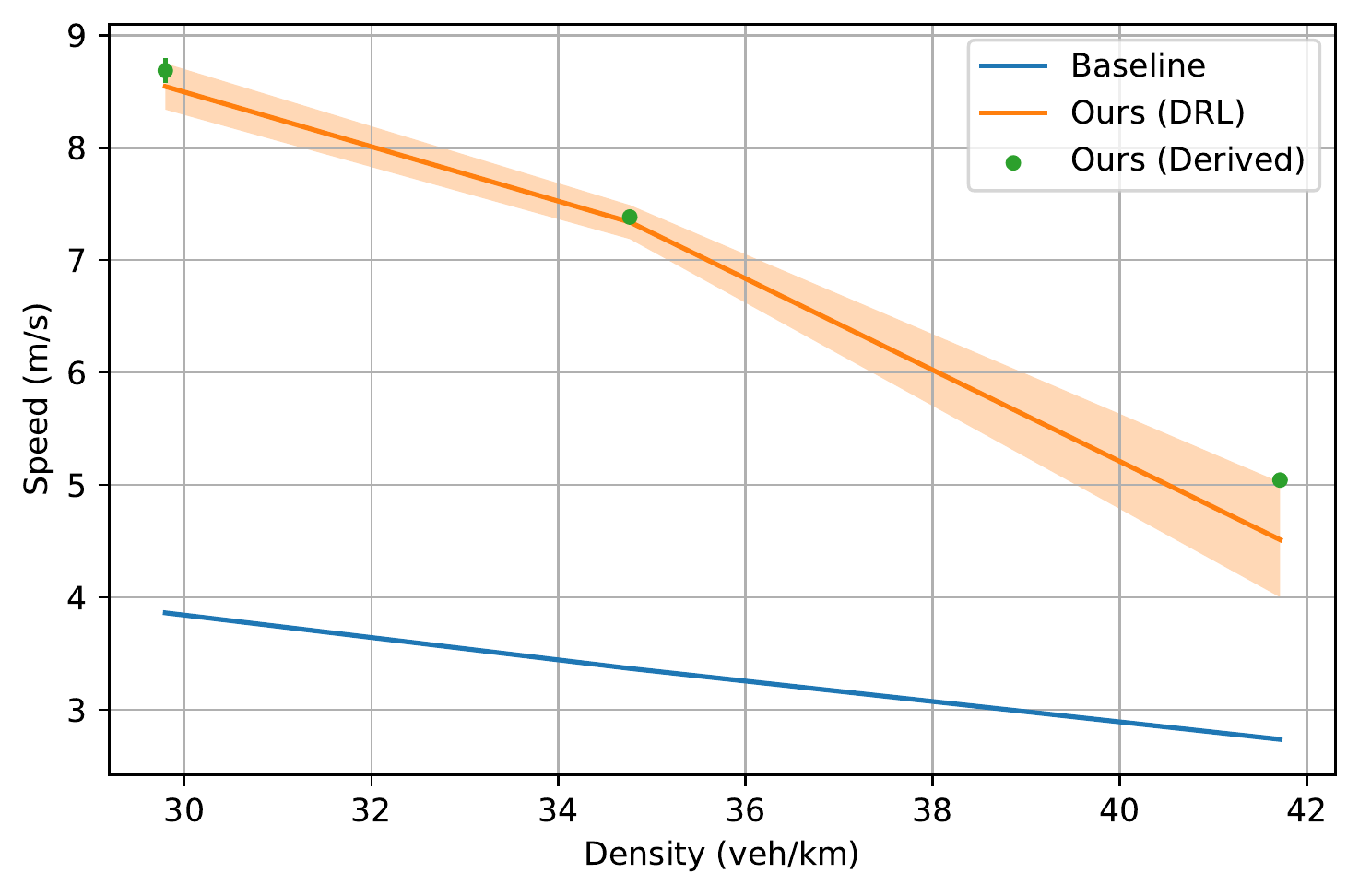}
\caption{\textbf{Figure Eight Average Speed.} We compare the average speed over all 14 vehicles over horizon $H$ under the Baseline, DRL, and Derived policies. The DRL policy nearly matches the Derived policy, despite needing to infer the target speed from solely the observations.}
\label{fig:figure_eight_speed}
\end{figure}
As the intersection is unsignalized, the Figure Eight system under the Baseline policy sees vehicles alternating to pass the intersection one by one, similar to the behavior at a stop-sign. As shown in Figure~\ref{fig:figure_eight_timespace}, the DRL policy instead learns to slow down to gather the rest of the vehicles as followers, then increases the speed while the other vehicles follow to ``snake'' around the Figure Eight. This behavior allows the speed of all vehicles to be faster than the average Baseline speed, as shown in Figure~\ref{fig:figure_eight_speed}. Using the same approach as the Single Ring, we design the Derived policy by applying exhaustive search to find an optimal target speed $v_\text{target}$. We find that DRL achieves close to the tuned target speed for all configurations.

\begin{algorithm}
\caption{Figure Eight Derived Policy}\label{alg:figure_eight_derived}
\begin{algorithmic}[0]
\Procedure{Derived}{$s$} \Comment{State $s$}
\State $R \leftarrow$ get radius from $s$
\State $x \leftarrow$ total distance of the figure eight
\State $x_\text{last} \leftarrow$ distance from the last follower to the AV
\If{$x_\text{last} < \frac{x}{2}$}
\State $v_\text{target} \leftarrow$ tuned target speed for $R$
\Else{} \Comment{Slow initial speed to gather followers}
\State $v_\text{target} \leftarrow 0.5$m/s
\EndIf
\State $v \leftarrow$ get speed of the AV from $s$
\State \Return Equalize($v_\text{target}, v$)
\EndProcedure
\end{algorithmic}
\end{algorithm}

While \cite{vinitsky2018benchmarks} reports similar DRL behavior in the Figure Eight systems, it shapes the reward function to explicitly encourages convergence towards a handpicked target speed. On the other hand, our present work demonstrates that simply optimizing for the end objective suffices without any handcrafting by the researcher or practitioner.

\subsection{Highway Bottleneck}
\label{sec:highway_bottleneck}
\begin{figure}[ht]
    \centering
    \subfigure[$F = 2000$~vehs/hr]{\includegraphics[width=3.5in]{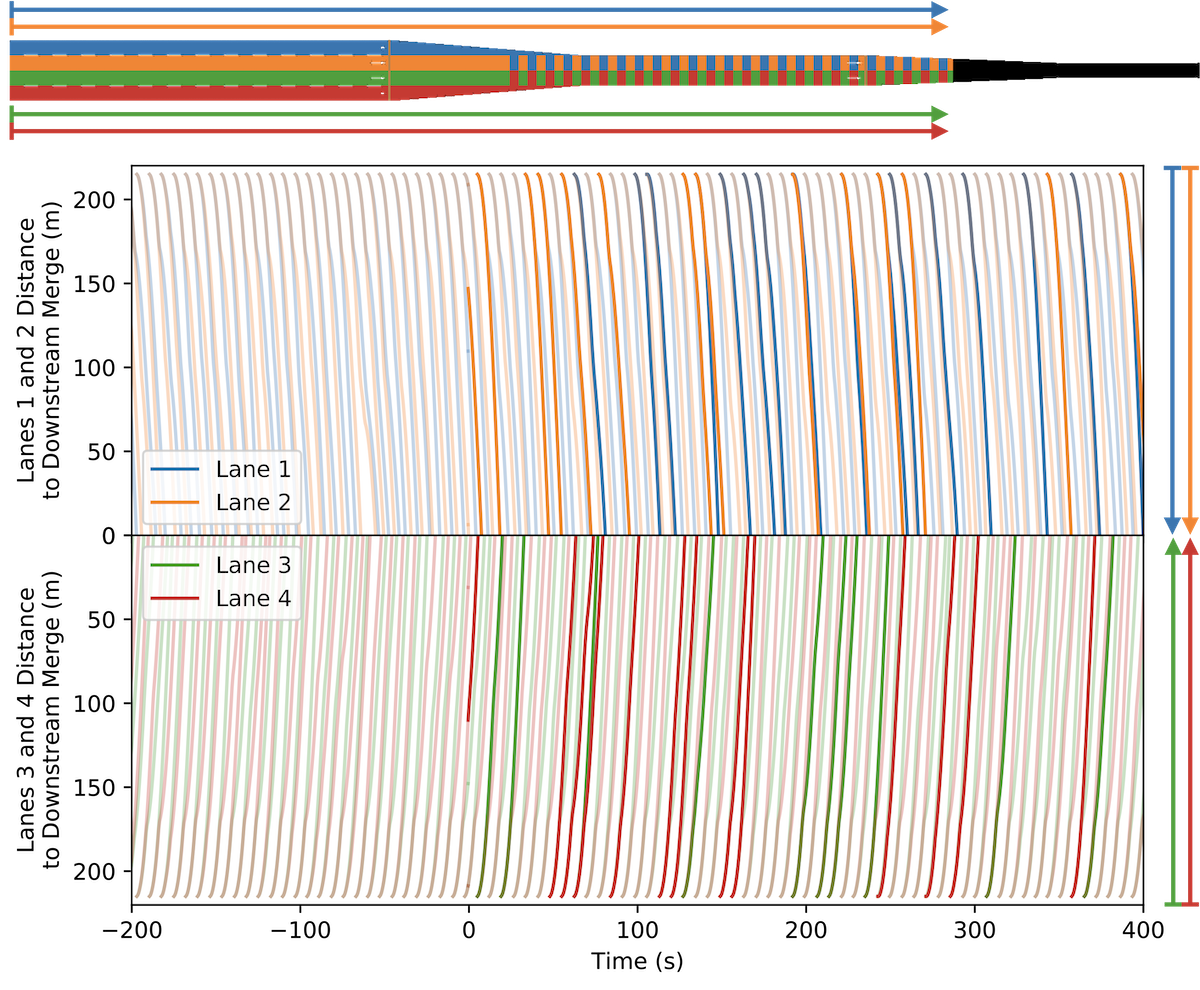}}
    \subfigure[$F = 2400$~vehs/hr]{\includegraphics[width=3.5in]{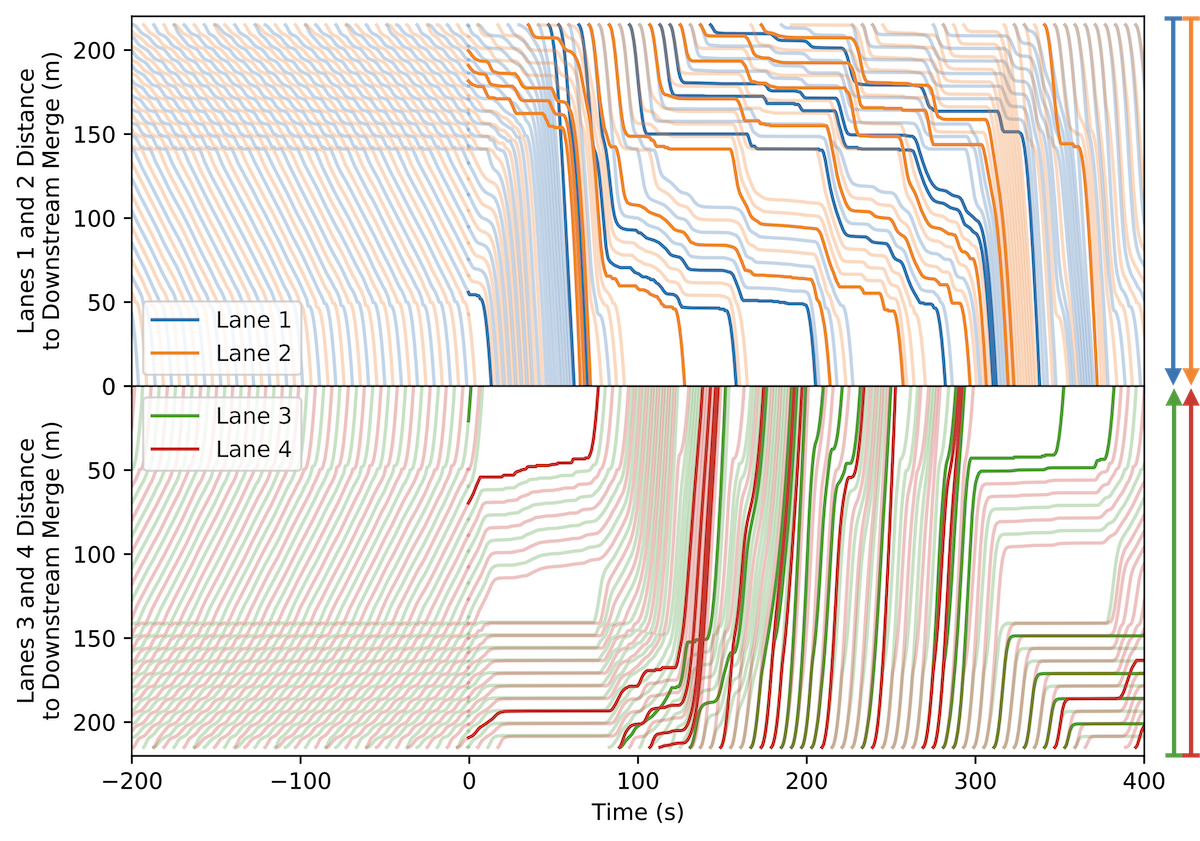}}
    \caption{\textbf{Highway Bottleneck Time-space Diagrams.} We plot the trajectories of vehicles under the Baseline policy (before time 0s) and the learned DRL policy (on and after time 0s). Blue, orange, green, and red lines indicate vehicles originating on lanes 1, 2, 3, and 4, respectively, and correspond to colored arrows indicating progressions of vehicles. Bold indicates the AVs controlled by the DRL policy. For $F < 2200$~vehs/hr, DRL sees the same efficient behavior as the Baseline. For $F \geq 2200$~vehs/hr, Baseline degrades significantly into an inefficient alternation, DRL reduces alternation by letting groups of vehicles pass the downstream bottleneck at once.}
    \label{fig:bottleneck_timespace}
\end{figure}

\begin{figure}[!t]
\centering
\includegraphics[width=3.5in]{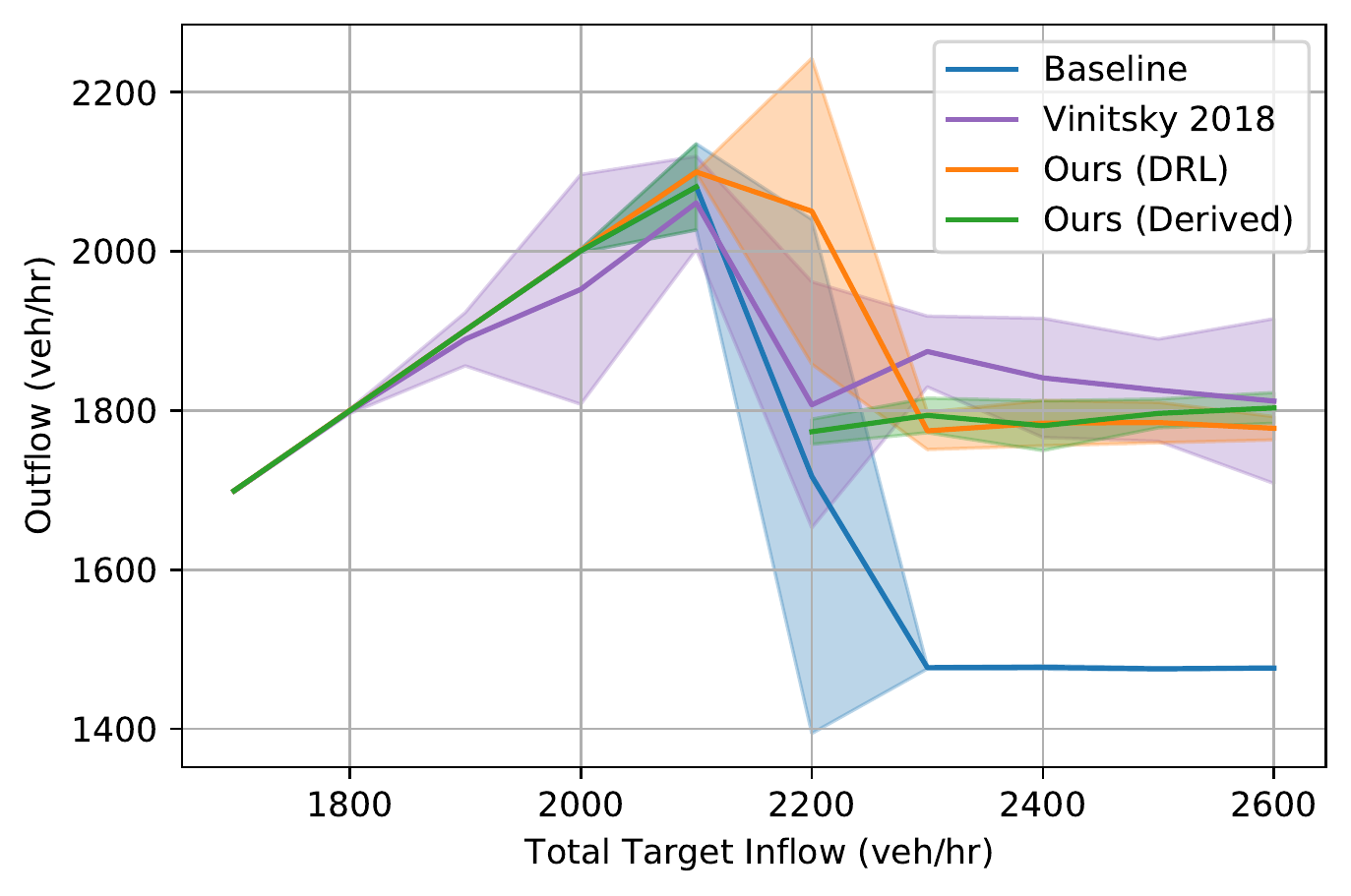}
\caption{\textbf{Highway Bottleneck Outflow.} We compare the outflow over horizon $H$ under the Baseline, DRL, and Derived policies\edit{, with the shaped deep reinforcement learning method Vinitsky 2018~\cite{vinitsky2018lagrangian} as additional comparison}. Our DRL policy sees similar performance to Derived under most target inflow rates, though \edit{the former} learns to mitigate the transition region ($F = 2200$~vehs/hr) better than \edit{the latter}. Both \edit{policies} are significantly better than Baseline at high target inflow rates. We visualize Derived as a piecewise function because Derived reverts to Baseline for $F \leq 2100$~vehs/hr and the optimal threshold parameters $x_1, x_2$ are shared for all $F \geq 2200$~vehs/hr. \edit{With better performance for $F \leq 2200$ and worse performance for $F \geq 2300$, our DRL policy performs similarly on average to Vinitsky 2018, which artificially restricts control of AVs to segments of the traffic system to encourage ramp metering-like behavior.}}
\label{fig:bottleneck_outflow}
\end{figure}

The Highway Bottleneck under the Baseline policy sees two distinct behaviors: at low target inflow rates $F < 2200$~vehs/hr, vehicles from the two merging lanes may weave together without slowing down; at high target inflow rates $F \geq 2200$~vehs/hr, a capacity drop phenomenon \cite{saberi2013hysteresis} occurs, and vehicles from the two merging lanes slow down to a near stop before the merge, taking turns to continue onto the merged lane. We observe that the behavior (Figure~\ref{fig:bottleneck_timespace}) of the trained DRL policy is similar to Baseline for $F < 2200$~vehs/hr; however, AVs learn to reduce alternation at merge points for $F \geq 2200$~vehs/hr, achieving higher throughput by letting a group of vehicles pass at once (Figure~\ref{fig:bottleneck_outflow}).

\edit{We consider the DRL method introduced by Vinitsky 2018~\cite{vinitsky2018lagrangian} as an additional baseline in Figure~\ref{fig:bottleneck_outflow}. While this prior work reduces the control space of the policy to upstream segments of the highway bottlenecks to encourage ramp metering behaviors, our methodology does not impose artificial restrictions to guide the policy. To train Vinitsky 2018, we augment the original method described by \cite{vinitsky2018lagrangian} with the RMSprop optimizer~\cite{hinton2012neural}, which we find to improve performance over the ADAM optimizer~\cite{kingma2014adam} used by \cite{vinitsky2018lagrangian}. Our DRL policy performs similarly on average to Vinitsky 2018, with better performance for lower $F$ and worse performance for higher $F$. These trade-offs in performance suggests that an interesting topic of future research may study the advantages and limitations of an unified methodology for segment-based control of mixed autonomy traffic.}
% While our numerical results resemble those of the ``Minimal'' policy reported by \cite{vinitsky2020optimizing} (differences in network geometry and vehicle characteristics mean that we cannot directly compare throughputs for each value of $F$), the prior work reduces the control space of the DRL policy to a narrow upstream region to encourage ramp metering behavior upstream from the merge point while we do not impose restrictions to guide the policy. Moreover, our policy exceeds or matches the Baseline performance for the entire range of $F$, while the policies in \cite{vinitsky2020optimizing} underperform the Baseline performance for an intermediate range of $F$.

For additional comparison, we design a Derived policy with tuned threshold parameters $x_1$ and $x_2$ which attempts to reduce alternation in a similar way to our policy if $F > 2200$~vehs/hr, otherwise mimicking Baseline behavior. Essentially, AV $i$ stops near the merge point if the following vehicle on the adjacent lane is uncontrolled and also near the merge point. This encourages AV $i$ to wait until the vehicle on the adjacent lane is an AV before continuing. The Derived policy suffers more at $F = 2200$~vehs/hr from the capacity drop but otherwise performs similarly to the DRL policy.
\begin{algorithm}
\caption{Highway Bottleneck Derived Policy}\label{alg:bottleneck_derived}
\begin{algorithmic}[0]
\Procedure{Derived}{$s$, $i$} \Comment{State $s$, AV index $i$}
\State $F \leftarrow$ get target inflow rate from $s$
\If{$F \leq 2200$}
\State \Return Uncontrolled($s, i$)
\EndIf
\State Let $j$ be the vehicle following $i$ in the adjacent lane
\State $x_1, x_2 \leftarrow$ tuned thresholds parameters
\State $d_i, d_j \leftarrow$ distances to the merge point for $i, j$
\State stop $\leftarrow$ $j$ is uncontrolled \textbf{and} $d_i < x_1$ \textbf{and} $d_j < x_2$
\State \Return $-c_\text{decel}$ \textbf{if} stop \textbf{else} $c_\text{accel}$
\EndProcedure
\\
\Procedure{Uncontrolled}{$s, i$}
\State \Return IDM acceleration for vehicle $i$ based on $s$
\EndProcedure
\end{algorithmic}
\end{algorithm}

\subsection{Highway Ramp}
\begin{figure}[ht]
    \centering
    \includegraphics[width=3.5in]{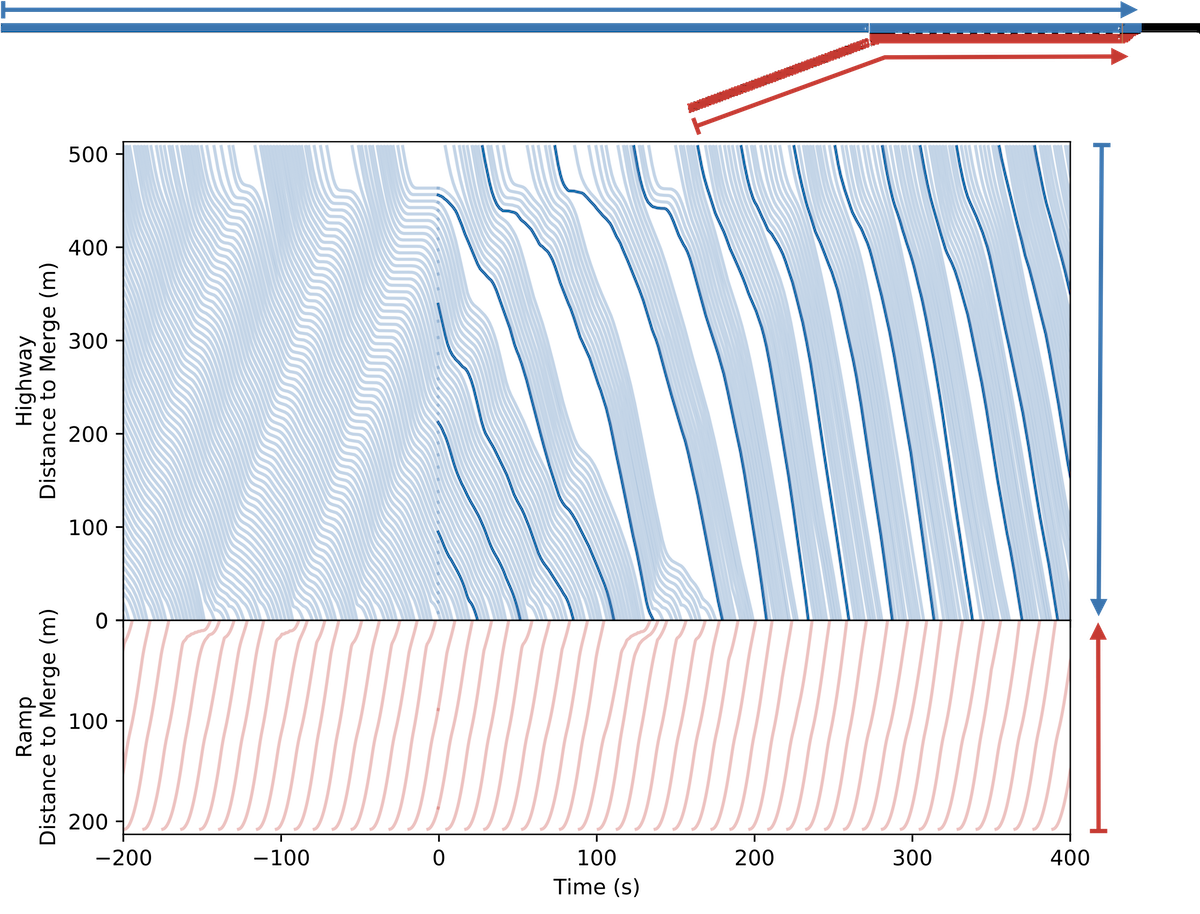}
    \caption{\textbf{Highway Ramp Time-space Diagrams.} We plot the trajectories of vehicles under the Baseline policy (before time 0s) and the learned DRL policy (on and after time 0s). Bold indicates the AVs controlled by the DRL policy. Colored arrows indicate progressions of highway and ramp vehicles approaching the merge. While vehicles slow down at the merge point in Baseline, DRL learns to regulate the upstream speed of the highway vehicles so that vehicles at the merge point do not slow down.}
    \label{fig:ramp_timespace}
\end{figure}

\begin{figure}[!t]
\centering
\includegraphics[width=3.5in]{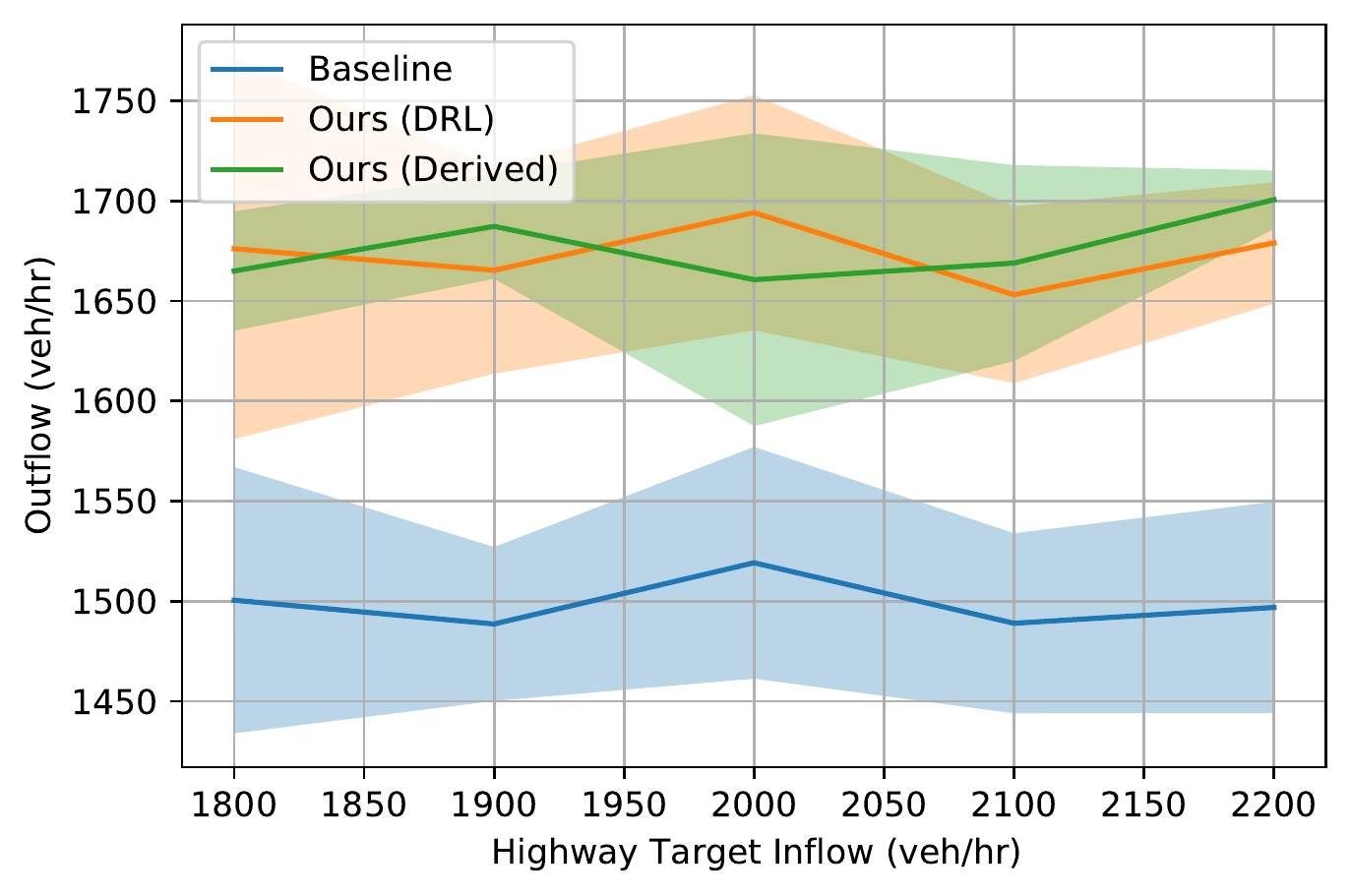}
\caption{\textbf{Highway Ramp Outflow.} We compare the outflow over horizon $H$ under the Baseline, DRL, and Derived policies. Derived and DRL perform similarly for all target inflow rates; unlike the Derived policy, the DRL policy is not informed of the congestion ahead of each AV and faces a more difficult task. We display Derived as a single curve because the same target speed parameter $v_\text{target}$ is optimal for all $F$ considered.}
\label{fig:ramp_outflow}
\end{figure}
In the Highway Ramp system under the Baseline policy, the ramp vehicles merging onto the highway force the highway vehicles to slow down, causing stop-and-go waves to propagate backward along the highway. The DRL policy learns to control AVs to hold highway vehicles back (Figure~\ref{fig:ramp_timespace}) to allow merging at a higher speed (Figure~\ref{fig:ramp_outflow}). The traffic system is similar to the one studied in \cite{kreidieh2018dissipating}, though we directly use the outflow as the objective while the prior work designs a reward function to encourage the speed of highway vehicle towards a manually specified $v_\text{des}$.

Observing the AV behavior under the DRL policy, we construct the Derived policy to similarly hold back highway vehicles distant from the merge point towards a tuned speed parameter $v_\text{target}$ to allow for higher speed at the merge point. If the highway ahead is congested, $v_\text{target}$ is temporarily set to $0$ to allow congestion to ease. The Derived policy performs similarly to the DRL policy but requires more information on the congestion in front of the AV, which is provided as $n_\text{leaders}$.

\begin{algorithm}
\caption{Highway Ramp Derived Policy}\label{alg:ramp_derived}
\begin{algorithmic}[0]
\Procedure{Derived}{$s$, $i$} \Comment{State $s$, AV index $i$}
\State $d_i \leftarrow$ distance to the merge point for AV $i$
\If{$d_i \leq 400$}
\State \Return Uncontrolled($s, i$)
\EndIf
\State $v_\text{target} \leftarrow$ tuned speed parameter
\State $v_i \leftarrow$ speed of AV $i$
\State $n_\text{leaders} \leftarrow$ number of vehicles in front of $i$
\If{$n_\text{leaders} > 20$} \Comment{Congested ahead}
\State $v_\text{target} \leftarrow 0$ \Comment{Wait for congestion to clear}
\EndIf
\State \Return Equalize($v_\text{target}, v_i$)
\EndProcedure
\end{algorithmic}
\end{algorithm}

\subsection{Intersection}
\begin{figure}[!t]
\centering
\includegraphics[width=3.5in]{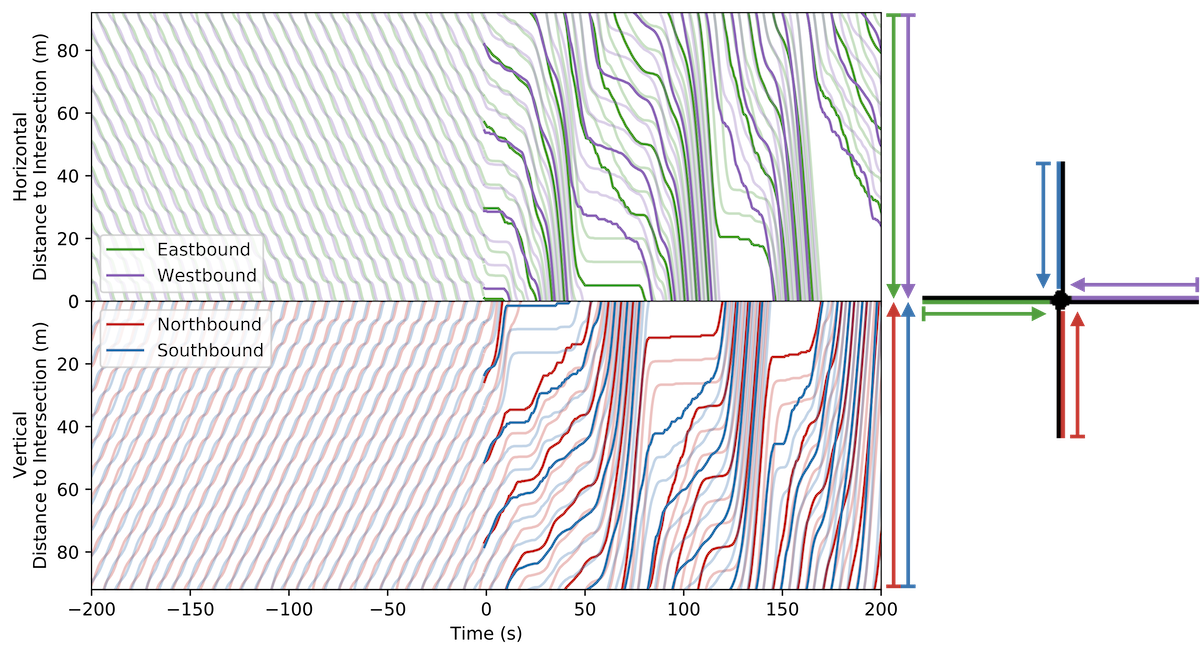}
\caption{\textbf{Intersection Time-space Diagrams.} We plot the trajectories of vehicles under the Baseline policy (before time 0s) and the learned DRL policy (on and after time 0s). Bold indicates the AVs controlled by the DRL policy. Colored arrows indicate progressions of vehicles on all lanes approaching the intersection. We see that the DRL policy develops an efficient traffic-signal-like behavior for grouping multiple vehicles and synchronizing the opposite lanes, whereas vehicles sees a stop-sign-like behavior under the Baseline policy.}
\label{fig:intersection_timespace}
\end{figure}

\begin{figure}[!t]
\centering
\includegraphics[width=3.5in]{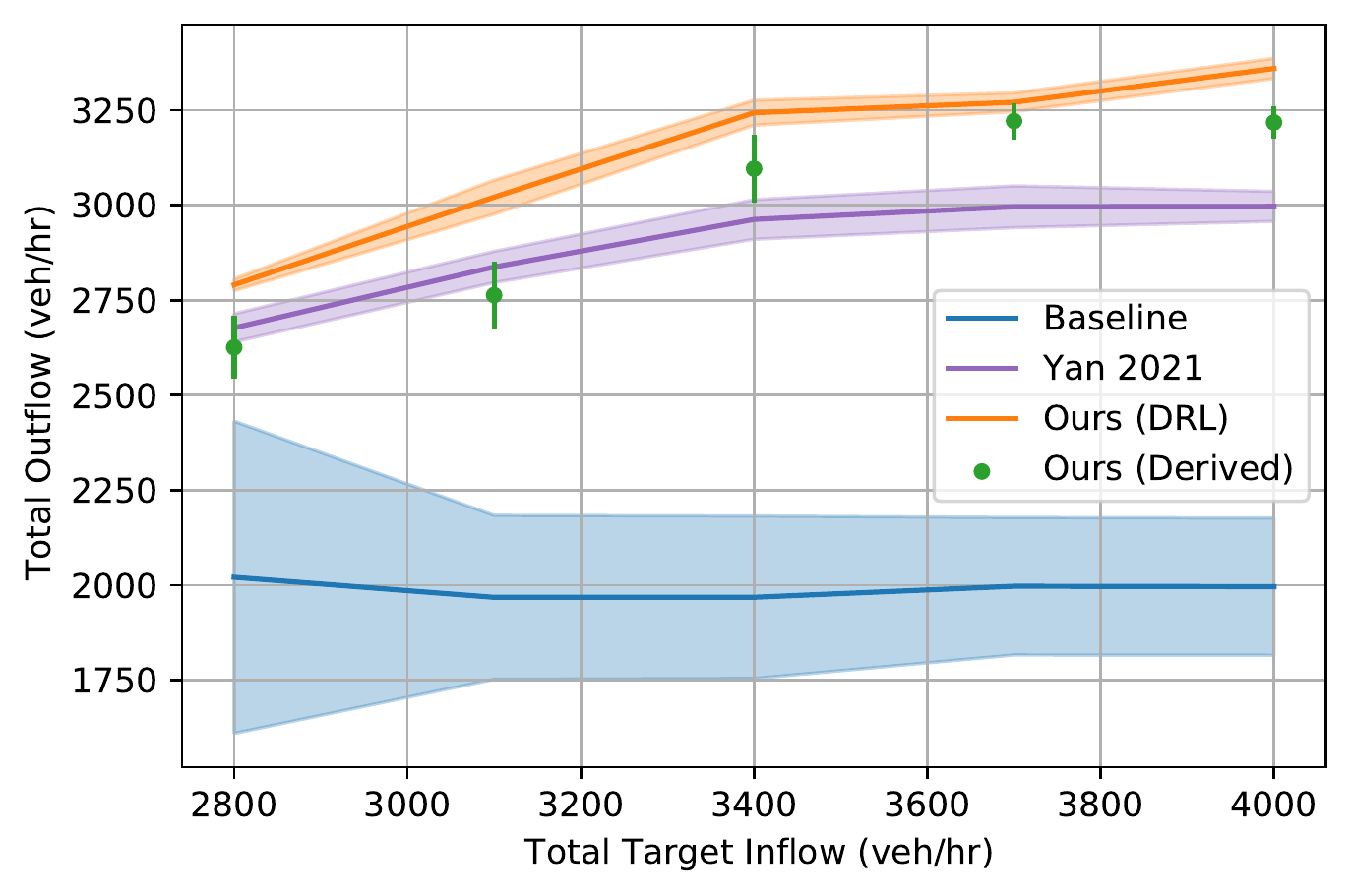}
\caption{\textbf{Intersection Outflow.} We compare the outflow over horizon $H$ under the Baseline, DRL, and Derived policies\edit{, with the deep reinforcement learning method Yan 2021~\cite{yan2021reinforcement} as additional comparison}. Note that performance is measured on all combinations of $(F_H, F_V) \in \{400, 550, 700, 850, 1000\}$~vehs/hr such that the total target inflow rate $F = 2(F_H + F_V)$ satisfies $2800 \text{~vehs/hr}\leq F \leq 4000$~vehs/hr. Though Derived attempts to mimic the traffic signal behavior of our DRL policy with tuned horizontal and vertical phases, we find it difficult to achieve DRL-level performance with handcrafting. This suggests that the DRL controller is both performant and robust across configurations of $F_H$ and $F_V$. \edit{Our DRL policy significantly outperforms Yan 2021 for all densities of traffic.}}
\label{fig:intersection_outflow}
\end{figure}

The Baseline Intersection system suffers severely from vehicles alternating to pass the intersection. DRL-controlled AVs not only learn to alternate less frequently, but they also learn to synchronize with AVs on opposite lanes (Figure~\ref{fig:intersection_timespace}). These learned behaviors resemble those of an adaptive traffic signal, greatly improving intersection throughput over the Baseline policy (Figure~\ref{fig:intersection_outflow}). Therefore, we design the Derived policy to follow a traffic signal-like behavior parameterized by horizontal and vertical phase $t_H$ and $t_V$, which are tuned for each density configuration, with no yellow time. The AV additionally yields to any uncontrolled vehicles currently crossing the intersection. Though $t_H$ and $t_V$ are tuned independently for each configuration, we find that the Derived policy suffers from occasional lapses into alternation. \edit{In an additional comparison to Yan 2021~\cite{yan2021reinforcement}, we demonstrate that our present learning rate-free TRPO-based methodology offers significant advantages over a REINFORCE-based methodology, which obtains worse performance even with careful tuning of the learning rate.}

\begin{algorithm}
\caption{Intersection Derived Policy}\label{alg:intersection_derived}
\begin{algorithmic}[0]
\Procedure{Derived}{$s$, $i$} \Comment{State $s$, AV index $i$}
\State $\ell_i, d_i \leftarrow$ lane, distance to intersection of AV $i$
\If{$d_i \geq 15$}
\State \Return Uncontrolled($s, i$)
\EndIf
\State $t_H, t_V \leftarrow$ tuned phase parameters
\State $t \leftarrow$ current simulation step $\mod (t_H + t_V)$
\State phase $\leftarrow$ horizontal \textbf{if} $t < t_H$ \textbf{else} vertical
\If{$\ell_i$ does not match phase}
\State \Return $-c_\text{decel}$
\ElsIf{uncontrolled vehicles are crossing}
\State \Return $-c_\text{decel}$
\Else{} \Return $c_\text{accel}$
\EndIf
\EndProcedure
\end{algorithmic}
\end{algorithm}

\section{Conclusion}
This article introduces a unified and straightforward methodology for optimizing vehicular systems with mixed or full autonomy. While we demonstrate the generality and effectiveness of our methodology on several mixed autonomy traffic systems, the same methodology could be adapted to other vehicular robotic systems \cite{wurman2008coordinating}. While our previous works applying DRL to mixed autonomy traffic often require extensive hyperparameter tuning and reward shaping, we show that the methodology presented in this work requires minimal hand-design and hyperparameter tuning. The performance and robustness of trained policies are characterized by comparisons with tuned rule-based policies. Finally, we provide future researchers and practitioners a lightweight framework which may be easily adapted to other systems and domains. While our earlier works based on the Flow framework \cite{wu2021flow,wu2017emergent,kreidieh2018dissipating,vinitsky2018benchmarks,vinitsky2020optimizing} were restricted by reliance on the general-use and heavyweight RLLib library \cite{liang2018rllib}, the unified methodology presented in this work is the result of our new lightweight framework which allowed more flexible research into the efficacy of various methodological components.

Further Sim2Real research and engineering are likely required for deployment in physical systems. \edit{In particular, future research may inject additional randomization and stochasticity in various aspects of the simulation to facilitate the learning of robust policies for Sim2Real transfer, while the design of the real-world system could be adjusted to reduce modeling error as much as possible.} We argue that near-future Sim2Real extensions of our work are feasible for automated systems with existence of high fidelity simulators and safety mechanisms, as long as human intent does not need to be simulated; these criteria are likely satisfied already in industrial robotic settings. \edit{As real-world deployment may benefit greatly from interpretability of trained policies, an impactful direction of future research may design an automatic method for distillation of trained DRL policies into Derived policies with interpretable behavioral building blocks, which could be shared across multiple systems. For example, convergence to fixed speed is a behavioral building block shared across DRL and Derived policies in Single Ring, Double Ring, and Figure Eight; waiting for a desired condition is another behavioral building block shared across Figure Eight, Highway Bottleneck, Highway Ramp, and Intersection.}

Open directions of research in vehicular systems and traffic systems include 1) application of our methodology to vehicular systems beyond traffic systems, 2) application to richer traffic systems with other maneuvers such as turning and other control elements such as traffic signals, 3) optimizing for non-efficiency objectives, such as fuel or comfort, 4) optimizing the behavior of heterogeneous vehicles with different physical properties, 5) systems with non-stationary traffic regimes (\textit{e.g.} natural variation of inflow rates), and 6) scaling up to larger systems by leveraging decomposition techniques. Due to multi-task training over many density configurations, we believe that our methodology already naturally handles non-stationary traffic regimes in particular, while the other open directions of research require further investigation.

\edit{\appendix[Robustness of Derived Policies under Ranges of Car Following Model Parameters]
\label{sec:appendix}
As uncontrolled vehicles in our simulated traffic systems follow the IDM~\cite{treiber2000congested} car following model, which models human driving with a set of behavioral parameters, simulation dynamics may differ under differing IDM parameters. To probe the robustness of our Derived policies in the context of other car following parameters, we consider the Highway Bottleneck at a total target inflow of $F = 2600$~vehs/hr. Here, we identify the default IDM parameters as maximum acceleration $a = 2.6$~m/s$^2$, comfortable deceleration $b = 4.5$~m/s$^2$, desired velocity $v_0 = 30$~m/s, minimum spacing $s_0 = 2.5$~m, desired time headway $\tau = 1$~s, and exponent $\delta = 4$. In Table~\ref{tab:idm_params}, we document the effect of reasonable changes in IDM parameters on the performance of the Derived and Baseline policies from Section~\ref{sec:highway_bottleneck}, holding the parameter of the Derived policy constant.}

\edit{Derived outperforms the Baseline for all parameter combinations, and the performance of both policies are positively correlated with $\delta$ and $v_0$ and negatively correlated with $\tau$ and $s_0$. The performance of Derived is positively correlated with $(a, b)$ while that of the Baseline is uncorrelated. In general, IDM parameter values modeling aggressive driving tend to improve the performance relative to those modeling conservative driving. Larger acceleration and deceleration parameters $(a, b)$, higher desired velocity $v_0$, smaller minimum spacing $s_0$ between vehicles, and smaller desired time headway $\tau$ all intuitively correspond to more aggressive driving, while smaller exponent $\delta$ increases the aggressiveness of accelerations when the vehicle speed is near $v_0$. The only exception is the lack of correlation between Baseline performance and reasonable ranges of $(a, b)$, suggesting that the inefficient alternation of vehicles at the bottleneck is not due to insufficient maximum acceleration $a$.}

\begin{table}[htp]

% \centering
\caption{\edit{Highway Bottleneck Outflow (vehs/hr) under different IDM parameters at total target inflow $F = 2600$~m/s. Each parameter table holds all other parameters at the default value. In all cases, means and standard deviations are computed over 10 independently sampled trajectories.}}\label{tab:idm_params}

\begin{tabular}{p{1cm}p{1.5cm}p{1.5cm}p{2.5cm}}
\toprule
$(a, b)$ & $(1, 1.5)$ &   $(2, 3)$ & $(2.6, 4.5)$ (Default) \\
\midrule
Baseline &   1476 $\pm$ 3 &   1476 $\pm$ 2 &               1476 $\pm$ 2 \\
Derived  &  1619 $\pm$ 27 &  1779 $\pm$ 24 &              1787 $\pm$ 26 \\
\bottomrule
\end{tabular}

\bigskip
\begin{tabular}{p{1cm}p{1.5cm}p{1.5cm}p{1.5cm}p{1.5cm}}
\toprule
$\tau$ &       $0.5$ &     $0.75$ & $1$ (Default) &     $1.25$ \\
\midrule
Baseline &  1696 $\pm$ 222 &   1540 $\pm$ 1 &      1476 $\pm$ 2 &   1456 $\pm$ 3 \\
Derived  &   2040 $\pm$ 30 &  1953 $\pm$ 19 &     1787 $\pm$ 26 &  1632 $\pm$ 15 \\
\bottomrule
\end{tabular}

\bigskip
\begin{tabular}{p{1cm}p{1.5cm}p{1.5cm}p{1.5cm}p{1.5cm}}
\toprule
$v_0$ &       $15$ &       $20$ &       $25$ & $30$ (Default) \\
\midrule
Baseline &   1463 $\pm$ 3 &   1474 $\pm$ 2 &   1475 $\pm$ 3 &       1476 $\pm$ 2 \\
Derived  &  1679 $\pm$ 13 &  1750 $\pm$ 31 &  1782 $\pm$ 20 &      1787 $\pm$ 26 \\
\bottomrule
\end{tabular}

\bigskip
\begin{tabular}{p{1cm}p{1.5cm}p{1.5cm}p{1.5cm}}
\toprule
$s_0$ &        $2$ & $2.5$ (Default) &        $3$ \\
\midrule
Baseline &   1514 $\pm$ 2 &        1476 $\pm$ 2 &   1440 $\pm$ 1 \\
Derived  &  1843 $\pm$ 28 &       1787 $\pm$ 26 &  1736 $\pm$ 17 \\
\bottomrule
\end{tabular}

\bigskip
\begin{tabular}{p{1cm}p{1.5cm}p{1.5cm}p{1.5cm}p{1.5cm}}
\toprule
$\delta$ &        $2$ &        $3$ & $4$ (Default) &        $5$ \\
\midrule
Baseline &   1458 $\pm$ 2 &   1473 $\pm$ 2 &      1476 $\pm$ 2 &   1478 $\pm$ 2 \\
Derived  &  1724 $\pm$ 17 &  1773 $\pm$ 26 &     1787 $\pm$ 26 &  1791 $\pm$ 24 \\
\bottomrule
\end{tabular}
\end{table}

\bibliographystyle{IEEEtran}
\bibliography{root}

\begin{thebibliography}{10}
\providecommand{\url}[1]{#1}
\csname url@samestyle\endcsname
\providecommand{\newblock}{\relax}
\providecommand{\bibinfo}[2]{#2}
\providecommand{\BIBentrySTDinterwordspacing}{\spaceskip=0pt\relax}
\providecommand{\BIBentryALTinterwordstretchfactor}{4}
\providecommand{\BIBentryALTinterwordspacing}{\spaceskip=\fontdimen2\font plus
\BIBentryALTinterwordstretchfactor\fontdimen3\font minus
  \fontdimen4\font\relax}
\providecommand{\BIBforeignlanguage}[2]{{%
\expandafter\ifx\csname l@#1\endcsname\relax
\typeout{** WARNING: IEEEtran.bst: No hyphenation pattern has been}%
\typeout{** loaded for the language `#1'. Using the pattern for}%
\typeout{** the default language instead.}%
\else
\language=\csname l@#1\endcsname
\fi
#2}}
\providecommand{\BIBdecl}{\relax}
\BIBdecl

\bibitem{morris2017guest}
K.~C. Morris, C.~Schlenoff, and V.~Srinivasan, ``Guest editorial a remarkable
  resurgence of artificial intelligence and its impact on automation and
  autonomy,'' \emph{IEEE Transactions on Automation Science and Engineering},
  vol.~14, no.~2, pp. 407--409, 2017.

\bibitem{wadud2016help}
Z.~Wadud, D.~MacKenzie, and P.~Leiby, ``Help or hindrance? the travel, energy
  and carbon impacts of highly automated vehicles,'' \emph{Transportation
  Research Part A: Policy and Practice}, vol.~86, pp. 1--18, 2016.

\bibitem{wurman2008coordinating}
P.~R. Wurman, R.~D'Andrea, and M.~Mountz, ``Coordinating hundreds of
  cooperative, autonomous vehicles in warehouses,'' \emph{AI magazine},
  vol.~29, no.~1, pp. 9--9, 2008.

\bibitem{wu2021flow}
C.~Wu, A.~Kreidieh, K.~Parvate, E.~Vinitsky, and A.~M. Bayen, ``Flow: A modular
  learning framework for mixed autonomy traffic,'' \emph{IEEE Transactions on
  Robotics}, 2021.

\bibitem{vinitsky2018lagrangian}
E.~Vinitsky, K.~Parvate, A.~Kreidieh, C.~Wu, and A.~Bayen, ``Lagrangian control
  through deep-rl: Applications to bottleneck decongestion,'' in \emph{2018
  21st International Conference on Intelligent Transportation Systems
  (ITSC)}.\hskip 1em plus 0.5em minus 0.4em\relax IEEE, 2018, pp. 759--765.

\bibitem{jang2019simulation}
K.~Jang, E.~Vinitsky, B.~Chalaki, B.~Remer, L.~Beaver, A.~A. Malikopoulos, and
  A.~Bayen, ``Simulation to scaled city: zero-shot policy transfer for traffic
  control via autonomous vehicles,'' in \emph{Proceedings of the 10th ACM/IEEE
  International Conference on Cyber-Physical Systems}, 2019, pp. 291--300.

\bibitem{hofer2021sim2real}
S.~H{\"o}fer, K.~Bekris, A.~Handa, J.~C. Gamboa, M.~Mozifian, F.~Golemo,
  C.~Atkeson, D.~Fox, K.~Goldberg, J.~Leonard \emph{et~al.}, ``Sim2real in
  robotics and automation: Applications and challenges,'' \emph{IEEE
  Transactions on Automation Science and Engineering}, vol.~18, no.~2, pp.
  398--400, 2021.

\bibitem{digani2015ensemble}
V.~Digani, L.~Sabattini, C.~Secchi, and C.~Fantuzzi, ``Ensemble coordination
  approach in multi-agv systems applied to industrial warehouses,'' \emph{IEEE
  Transactions on Automation Science and Engineering}, vol.~12, no.~3, pp.
  922--934, 2015.

\bibitem{lin2011optimization}
W.-S. Lin and J.-W. Sheu, ``Optimization of train regulation and energy usage
  of metro lines using an adaptive-optimal-control algorithm,'' \emph{IEEE
  Transactions on Automation Science and Engineering}, vol.~8, no.~4, pp.
  855--864, 2011.

\bibitem{morris2016planning}
R.~Morris, C.~S. Pasareanu, K.~Luckow, W.~Malik, H.~Ma, T.~S. Kumar, and
  S.~Koenig, ``Planning, scheduling and monitoring for airport surface
  operations,'' in \emph{Workshops at the Thirtieth AAAI Conference on
  Artificial Intelligence}, 2016.

\bibitem{stavrou2017optimizing}
D.~Stavrou, S.~Timotheou, C.~G. Panayiotou, and M.~M. Polycarpou, ``Optimizing
  container loading with autonomous robots,'' \emph{IEEE Transactions on
  Automation Science and Engineering}, vol.~15, no.~2, pp. 717--731, 2017.

\bibitem{zhu2020optimal}
Y.~Zhu, D.~Zhao, and H.~He, ``Optimal feedback control of pedestrian flow in
  heterogeneous corridors,'' \emph{IEEE Transactions on Automation Science and
  Engineering}, 2020.

\bibitem{wu2017emergent}
C.~Wu, A.~Kreidieh, E.~Vinitsky, and A.~M. Bayen, ``Emergent behaviors in
  mixed-autonomy traffic,'' in \emph{Conference on Robot Learning}.\hskip 1em
  plus 0.5em minus 0.4em\relax PMLR, 2017, pp. 398--407.

\bibitem{kreidieh2018dissipating}
A.~R. Kreidieh, C.~Wu, and A.~M. Bayen, ``Dissipating stop-and-go waves in
  closed and open networks via deep reinforcement learning,'' in \emph{2018
  21st International Conference on Intelligent Transportation Systems
  (ITSC)}.\hskip 1em plus 0.5em minus 0.4em\relax IEEE, 2018, pp. 1475--1480.

\bibitem{vinitsky2018benchmarks}
E.~Vinitsky, A.~Kreidieh, L.~Le~Flem, N.~Kheterpal, K.~Jang, C.~Wu, F.~Wu,
  R.~Liaw, E.~Liang, and A.~M. Bayen, ``Benchmarks for reinforcement learning
  in mixed-autonomy traffic,'' in \emph{Conference on robot learning}.\hskip
  1em plus 0.5em minus 0.4em\relax PMLR, 2018, pp. 399--409.

\bibitem{vinitsky2020optimizing}
E.~Vinitsky, N.~Lichtle, K.~Parvate, and A.~Bayen, ``Optimizing mixed autonomy
  traffic flow with decentralized autonomous vehicles and multi-agent rl,''
  \emph{arXiv preprint arXiv:2011.00120}, 2020.

\bibitem{yan2021reinforcement}
Z.~Yan and C.~Wu, ``Reinforcement learning for mixed autonomy intersections,''
  in \emph{2021 IEEE International Intelligent Transportation Systems
  Conference (ITSC)}.\hskip 1em plus 0.5em minus 0.4em\relax IEEE, 2021, pp.
  2089--2094.

\bibitem{papageorgiou2003review}
M.~Papageorgiou, C.~Diakaki, V.~Dinopoulou, A.~Kotsialos, and Y.~Wang, ``Review
  of road traffic control strategies,'' \emph{Proceedings of the IEEE},
  vol.~91, no.~12, pp. 2043--2067, 2003.

\bibitem{little1981maxband}
J.~D. Little, M.~D. Kelson, and N.~H. Gartner, ``Maxband: A versatile program
  for setting signals on arteries and triangular networks,'' 1981.

\bibitem{hunt1981scoot}
P.~Hunt, D.~Robertson, R.~Bretherton, R.~Winton, Transport, and R.~R.
  Laboratory, \emph{SCOOT: A Traffic Responsive Method of Coordinating
  Signals}, ser. TRRL Laboratory report.\hskip 1em plus 0.5em minus 0.4em\relax
  TRRL Urban Networks Division, 1981.

\bibitem{treiber2013traffic}
M.~Treiber and A.~Kesting, ``Traffic flow dynamics,'' \emph{Traffic Flow
  Dynamics: Data, Models and Simulation, Springer-Verlag Berlin Heidelberg},
  2013.

\bibitem{papageorgiou1991alinea}
M.~Papageorgiou, H.~Hadj-Salem, J.-M. Blosseville \emph{et~al.}, ``Alinea: A
  local feedback control law for on-ramp metering,'' \emph{Transportation
  research record}, vol. 1320, no.~1, pp. 58--67, 1991.

\bibitem{van2006impact}
B.~Van~Arem, C.~J. Van~Driel, and R.~Visser, ``The impact of cooperative
  adaptive cruise control on traffic-flow characteristics,'' \emph{IEEE
  Transactions on intelligent transportation systems}, vol.~7, no.~4, pp.
  429--436, 2006.

\bibitem{buehler2009darpa}
M.~Buehler, K.~Iagnemma, and S.~Singh, \emph{The DARPA urban challenge:
  autonomous vehicles in city traffic}.\hskip 1em plus 0.5em minus 0.4em\relax
  springer, 2009, vol.~56.

\bibitem{vahidi2003research}
A.~Vahidi and A.~Eskandarian, ``Research advances in intelligent collision
  avoidance and adaptive cruise control,'' \emph{IEEE transactions on
  intelligent transportation systems}, vol.~4, no.~3, pp. 143--153, 2003.

\bibitem{lefevre2015learning}
S.~Lefevre, A.~Carvalho, and F.~Borrelli, ``A learning-based framework for
  velocity control in autonomous driving,'' \emph{IEEE Transactions on
  Automation Science and Engineering}, vol.~13, no.~1, pp. 32--42, 2015.

\bibitem{sharon2017protocol}
G.~Sharon and P.~Stone, ``A protocol for mixed autonomous and human-operated
  vehicles at intersections,'' in \emph{International Conference on Autonomous
  Agents and Multiagent Systems}.\hskip 1em plus 0.5em minus 0.4em\relax
  Springer, 2017, pp. 151--167.

\bibitem{dresner2008multiagent}
K.~Dresner and P.~Stone, ``A multiagent approach to autonomous intersection
  management,'' \emph{Journal of artificial intelligence research}, vol.~31,
  pp. 591--656, 2008.

\bibitem{wu2012cooperative}
J.~Wu, A.~Abbas-Turki, and A.~El~Moudni, ``Cooperative driving: an ant colony
  system for autonomous intersection management,'' \emph{Applied Intelligence},
  vol.~37, no.~2, pp. 207--222, 2012.

\bibitem{miculescu2019polling}
D.~Miculescu and S.~Karaman, ``Polling-systems-based autonomous vehicle
  coordination in traffic intersections with no traffic signals,'' \emph{IEEE
  Transactions on Automatic Control}, vol.~65, no.~2, pp. 680--694, 2019.

\bibitem{wang2020thirty}
J.~Wang, C.~Jiang, H.~Zhang, Y.~Ren, K.-C. Chen, and L.~Hanzo, ``Thirty years
  of machine learning: The road to pareto-optimal wireless networks,''
  \emph{IEEE Communications Surveys \& Tutorials}, vol.~22, no.~3, pp.
  1472--1514, 2020.

\bibitem{ko2018wireless}
E.~Ko and K.-C. Chen, ``Wireless communications meets artificial intelligence:
  An illustration by autonomous vehicles on manhattan streets,'' in \emph{2018
  IEEE Global Communications Conference (GLOBECOM)}.\hskip 1em plus 0.5em minus
  0.4em\relax IEEE, 2018, pp. 1--7.

\bibitem{bertsekas1996neuro}
D.~P. Bertsekas and J.~N. Tsitsiklis, \emph{Neuro-dynamic programming}.\hskip
  1em plus 0.5em minus 0.4em\relax Athena Scientific, 1996.

\bibitem{sutton2018reinforcement}
R.~S. Sutton and A.~G. Barto, \emph{Reinforcement learning: An
  introduction}.\hskip 1em plus 0.5em minus 0.4em\relax MIT press, 2018.

\bibitem{wei2019colight}
H.~Wei, N.~Xu, H.~Zhang, G.~Zheng, X.~Zang, C.~Chen, W.~Zhang, Y.~Zhu, K.~Xu,
  and Z.~Li, ``Colight: Learning network-level cooperation for traffic signal
  control,'' in \emph{Proceedings of the 28th ACM International Conference on
  Information and Knowledge Management}, 2019, pp. 1913--1922.

\bibitem{belletti2017expert}
F.~Belletti, D.~Haziza, G.~Gomes, and A.~M. Bayen, ``Expert level control of
  ramp metering based on multi-task deep reinforcement learning,'' \emph{IEEE
  Transactions on Intelligent Transportation Systems}, vol.~19, no.~4, pp.
  1198--1207, 2017.

\bibitem{liu2021visuomotor}
Z.~Liu, Q.~Liu, L.~Tang, K.~Jin, H.~Wang, M.~Liu, and H.~Wang, ``Visuomotor
  reinforcement learning for multirobot cooperative navigation,'' \emph{IEEE
  Transactions on Automation Science and Engineering}, 2021.

\bibitem{xiao2019meta}
Q.~Xiao, C.~Li, Y.~Tang, and L.~Li, ``Meta-reinforcement learning of machining
  parameters for energy-efficient process control of flexible turning
  operations,'' \emph{IEEE Transactions on Automation Science and Engineering},
  2019.

\bibitem{park2019reinforcement}
I.-B. Park, J.~Huh, J.~Kim, and J.~Park, ``A reinforcement learning approach to
  robust scheduling of semiconductor manufacturing facilities,'' \emph{IEEE
  Transactions on Automation Science and Engineering}, vol.~17, no.~3, pp.
  1420--1431, 2019.

\bibitem{ou2020method}
X.~Ou, Q.~Chang, and N.~Chakraborty, ``A method integrating q-learning with
  approximate dynamic programming for gantry work cell scheduling,'' \emph{IEEE
  Transactions on Automation Science and Engineering}, vol.~18, no.~1, pp.
  85--93, 2020.

\bibitem{wu2017framework}
C.~Wu, K.~Parvate, N.~Kheterpal, L.~Dickstein, A.~Mehta, E.~Vinitsky, and A.~M.
  Bayen, ``Framework for control and deep reinforcement learning in traffic,''
  in \emph{2017 IEEE 20th International Conference on Intelligent
  Transportation Systems (ITSC)}.\hskip 1em plus 0.5em minus 0.4em\relax IEEE,
  2017, pp. 1--8.

\bibitem{schulman2015trust}
J.~Schulman, S.~Levine, P.~Abbeel, M.~Jordan, and P.~Moritz, ``Trust region
  policy optimization,'' in \emph{International conference on machine
  learning}.\hskip 1em plus 0.5em minus 0.4em\relax PMLR, 2015, pp. 1889--1897.

\bibitem{fujimoto2018addressing}
S.~Fujimoto, H.~Hoof, and D.~Meger, ``Addressing function approximation error
  in actor-critic methods,'' in \emph{International Conference on Machine
  Learning}.\hskip 1em plus 0.5em minus 0.4em\relax PMLR, 2018, pp. 1587--1596.

\bibitem{williams1992simple}
R.~J. Williams, ``Simple statistical gradient-following algorithms for
  connectionist reinforcement learning,'' \emph{Machine learning}, vol.~8, no.
  3-4, pp. 229--256, 1992.

\bibitem{kingma2014adam}
D.~P. Kingma and J.~Ba, ``Adam: {A} method for stochastic optimization,'' in
  \emph{3rd International Conference on Learning Representations, {ICLR} 2015,
  San Diego, CA, USA, May 7-9, 2015, Conference Track Proceedings}, 2015.

\bibitem{hinton2012neural}
G.~Hinton, N.~Srivastava, and K.~Swersky, ``Neural networks for machine
  learning lecture 6a overview of mini-batch gradient descent,'' \emph{Cited
  on}, vol.~14, no.~8, p.~2, 2012.

\bibitem{schulman2016high}
J.~Schulman, P.~Moritz, S.~Levine, M.~I. Jordan, and P.~Abbeel,
  ``High-dimensional continuous control using generalized advantage
  estimation,'' in \emph{4th International Conference on Learning
  Representations, {ICLR} 2016, San Juan, Puerto Rico, May 2-4, 2016,
  Conference Track Proceedings}, 2016.

\bibitem{treiber2000congested}
M.~Treiber, A.~Hennecke, and D.~Helbing, ``Congested traffic states in
  empirical observations and microscopic simulations,'' \emph{Physical review
  E}, vol.~62, no.~2, p. 1805, 2000.

\bibitem{liu2020prediction}
Z.~Liu, H.~Wang, H.~Wei, M.~Liu, and Y.-H. Liu, ``Prediction, planning, and
  coordination of thousand-warehousing-robot networks with motion and
  communication uncertainties,'' \emph{IEEE Transactions on Automation Science
  and Engineering}, 2020.

\bibitem{kaelbling1998planning}
L.~P. Kaelbling, M.~L. Littman, and A.~R. Cassandra, ``Planning and acting in
  partially observable stochastic domains,'' \emph{Artificial intelligence},
  vol. 101, no. 1-2, pp. 99--134, 1998.

\bibitem{boutilier1996planning}
C.~Boutilier, ``Planning, learning and coordination in multiagent decision
  processes,'' in \emph{TARK}, vol.~96.\hskip 1em plus 0.5em minus 0.4em\relax
  Citeseer, 1996, pp. 195--210.

\bibitem{gupta2017cooperative}
J.~K. Gupta, M.~Egorov, and M.~Kochenderfer, ``Cooperative multi-agent control
  using deep reinforcement learning,'' in \emph{International Conference on
  Autonomous Agents and Multiagent Systems}.\hskip 1em plus 0.5em minus
  0.4em\relax Springer, 2017, pp. 66--83.

\bibitem{engstrom2019implementation}
L.~Engstrom, A.~Ilyas, S.~Santurkar, D.~Tsipras, F.~Janoos, L.~Rudolph, and
  A.~Madry, ``Implementation matters in deep rl: A case study on ppo and
  trpo,'' in \emph{International conference on learning representations}, 2019.

\bibitem{kakade2001natural}
S.~M. Kakade, ``A natural policy gradient,'' \emph{Advances in neural
  information processing systems}, vol.~14, 2001.

\bibitem{agarwal2021theory}
A.~Agarwal, S.~M. Kakade, J.~D. Lee, and G.~Mahajan, ``On the theory of policy
  gradient methods: Optimality, approximation, and distribution shift,''
  \emph{Journal of Machine Learning Research}, vol.~22, no.~98, pp. 1--76,
  2021.

\bibitem{lopez2018microscopic}
P.~A. Lopez, M.~Behrisch, L.~Bieker-Walz, J.~Erdmann, Y.-P. Fl{\"o}tter{\"o}d,
  R.~Hilbrich, L.~L{\"u}cken, J.~Rummel, P.~Wagner, and E.~Wie{\ss}ner,
  ``Microscopic traffic simulation using sumo,'' in \emph{2018 21st
  International Conference on Intelligent Transportation Systems (ITSC)}.\hskip
  1em plus 0.5em minus 0.4em\relax IEEE, 2018, pp. 2575--2582.

\bibitem{herman1959traffic}
R.~Herman, E.~W. Montroll, R.~B. Potts, and R.~W. Rothery, ``Traffic dynamics:
  analysis of stability in car following,'' \emph{Operations research}, vol.~7,
  no.~1, pp. 86--106, 1959.

\bibitem{stern2018dissipation}
R.~E. Stern, S.~Cui, M.~L. Delle~Monache, R.~Bhadani, M.~Bunting, M.~Churchill,
  N.~Hamilton, H.~Pohlmann, F.~Wu, B.~Piccoli \emph{et~al.}, ``Dissipation of
  stop-and-go waves via control of autonomous vehicles: Field experiments,''
  \emph{Transportation Research Part C: Emerging Technologies}, vol.~89, pp.
  205--221, 2018.

\bibitem{saberi2013hysteresis}
M.~Saberi and H.~S. Mahmassani, ``Hysteresis and capacity drop phenomena in
  freeway networks: empirical characterization and interpretation,''
  \emph{Transportation research record}, vol. 2391, no.~1, pp. 44--55, 2013.

\bibitem{liang2018rllib}
E.~Liang, R.~Liaw, R.~Nishihara, P.~Moritz, R.~Fox, K.~Goldberg, J.~Gonzalez,
  M.~Jordan, and I.~Stoica, ``Rllib: Abstractions for distributed reinforcement
  learning,'' in \emph{International Conference on Machine Learning}.\hskip 1em
  plus 0.5em minus 0.4em\relax PMLR, 2018, pp. 3053--3062.

\end{thebibliography}

\begin{IEEEbiography}[{\includegraphics[width=1in,height=1.25in,clip,keepaspectratio]{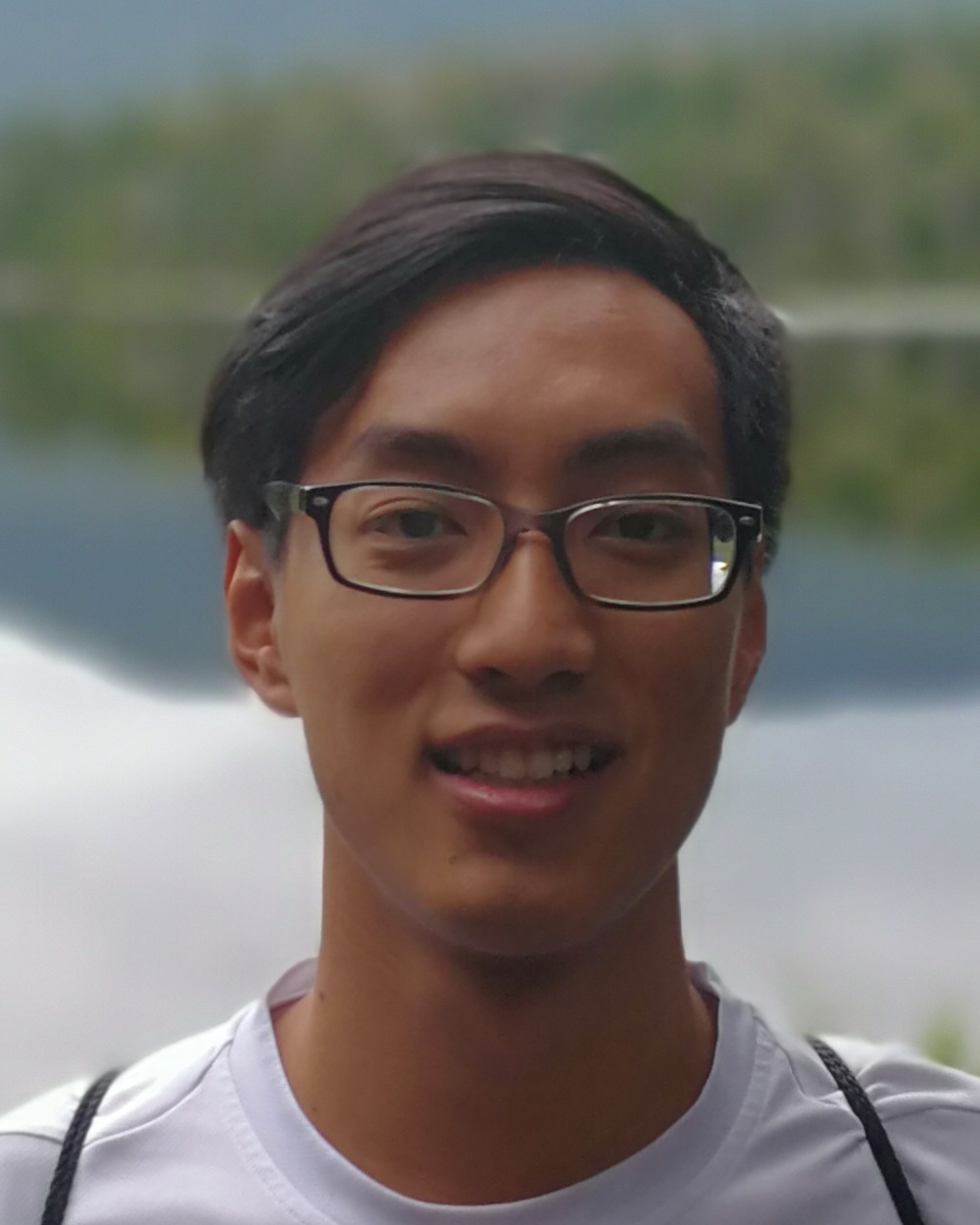}}]{Zhongxia Yan}
received the B.S. and M.S. degrees in electrical engineering and computer science from the University of California, Berkeley, Berkeley, CA, USA, in 2017 and 2018, respectively. He is currently working toward the Ph.D. degree in electrical engineering and computer science at Massachusetts Institute of Technology, Cambridge, MA, USA. His research interests include application of machine learning and reinforcement learning to problems in transportation and logistics. Mr. Yan is a recipient of the Department of Transportation DDETFP Graduate Fellowship.
\end{IEEEbiography}

\begin{IEEEbiography}[{\includegraphics[width=1in,height=1.25in,clip,keepaspectratio]{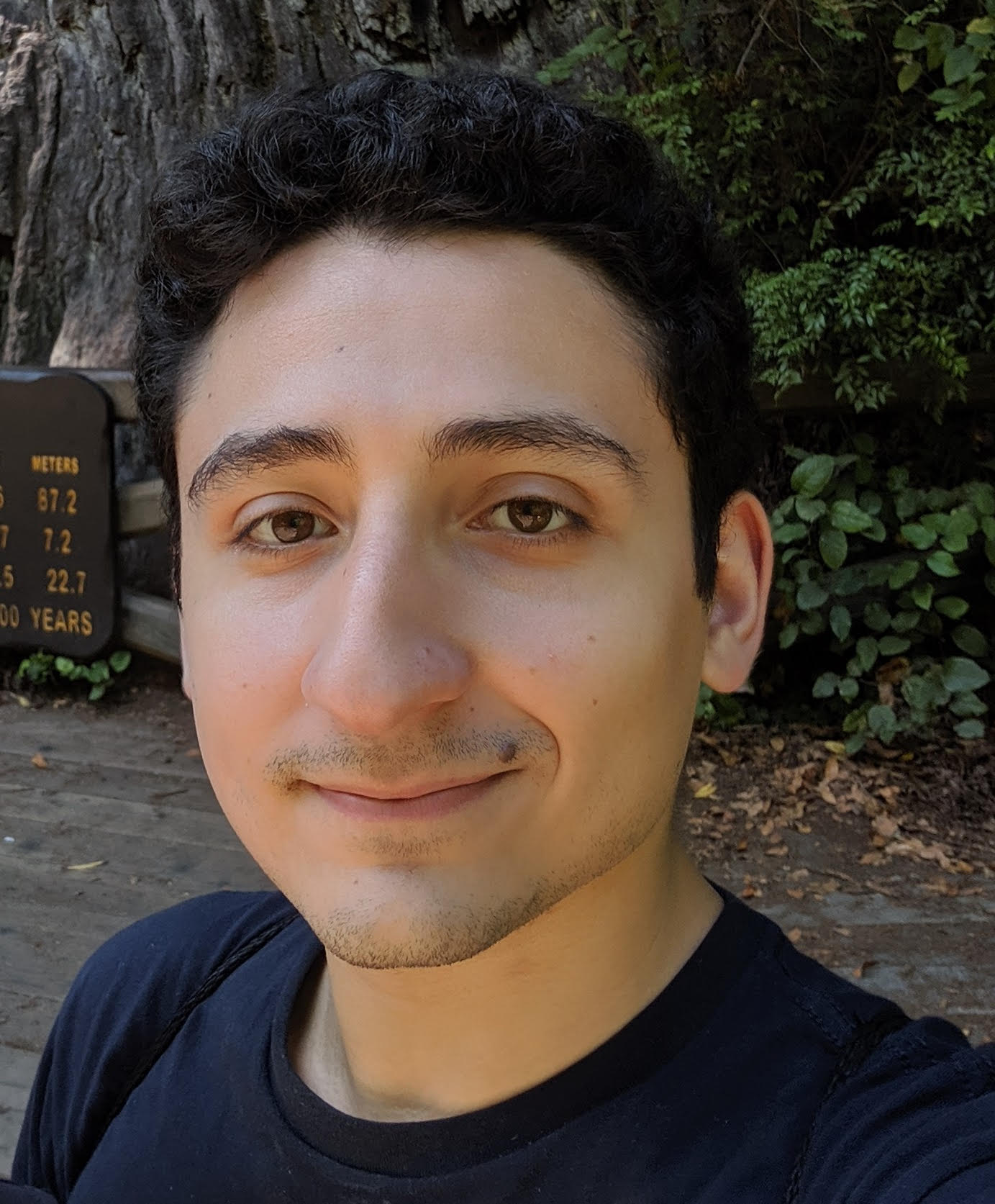}}]{Abdul Rahman Kreidieh}
received the B.S. degree in mechanical engineering from the American University of Beirut, Beirut, Lebanon, in 2016, and the M.S. degree in civil and environmental engineering in 2017 from the University of California, Berkeley, Berkeley, CA, USA, where he is currently working toward the Ph.D. degree in civil and environmental engineering. His primary focus is on designing models and algorithms that scale the performance of existing machine learning systems to large-scale traffic control problems. His research interests include the intersection of machine learning and traffic control through mixed autonomy systems.
\end{IEEEbiography}

\begin{IEEEbiography}[{\includegraphics[width=1in,height=1.25in,clip,keepaspectratio]{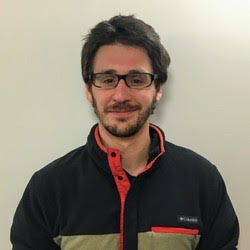}}]{Eugene Vinitsky}
received the B.S. degree in physics from the California Institute of Technology, Pasadena, CA, USA, in 2014, the M.A. degree in physics from the University of California, Santa Barbara, Santa Barbara, CA, in 2015. He is currently working toward the Ph.D. degree in controls engineering with the Mobile Sensing Laboratory, University of California, Berkeley, Berkeley, CA. His current research interests include multi-agent reinforcement learning, cooperative automated vehicles, and robust control. Mr. Vinitsky is a recipient of a National Science Foundation Graduate Research Fellowship.
\end{IEEEbiography}

\begin{IEEEbiography}[{\includegraphics[width=1in,height=1.25in,clip,keepaspectratio]{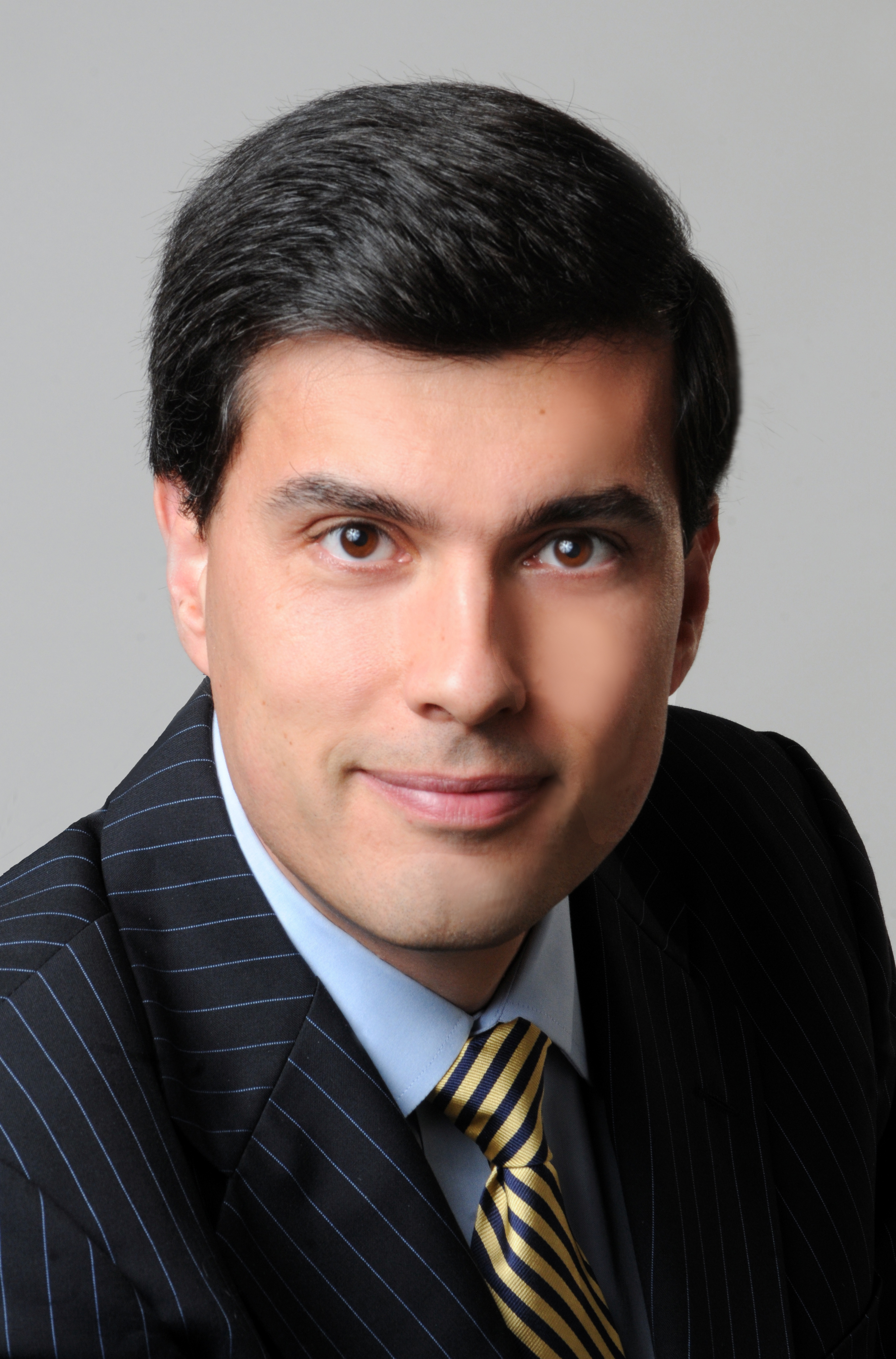}}]{Alexandre M. Bayen}
received the Engineering degree in applied mathematics from the Ecole Polytechnique, Palaiseau, France, in 1998, and the M.S. and Ph.D. degrees in aeronautics and astronautics from Stanford University, Stanford, CA, USA, in 1999 and 2004, respectively. He is the Liao-Cho Professor of Engineering with University of California, Berkeley, Berkeley, CA, USA. He is currently a Professor of Electrical Engineering and Computer Science, and Civil and Environmental Engineering. He is currently the Director of the Institute of Transportation Studies. He is also a Faculty Scientist in Mechanical Engineering, Lawrence Berkeley National Laboratory.
\end{IEEEbiography}

\begin{IEEEbiography}[{\includegraphics[width=1in,height=1.25in,clip,keepaspectratio]{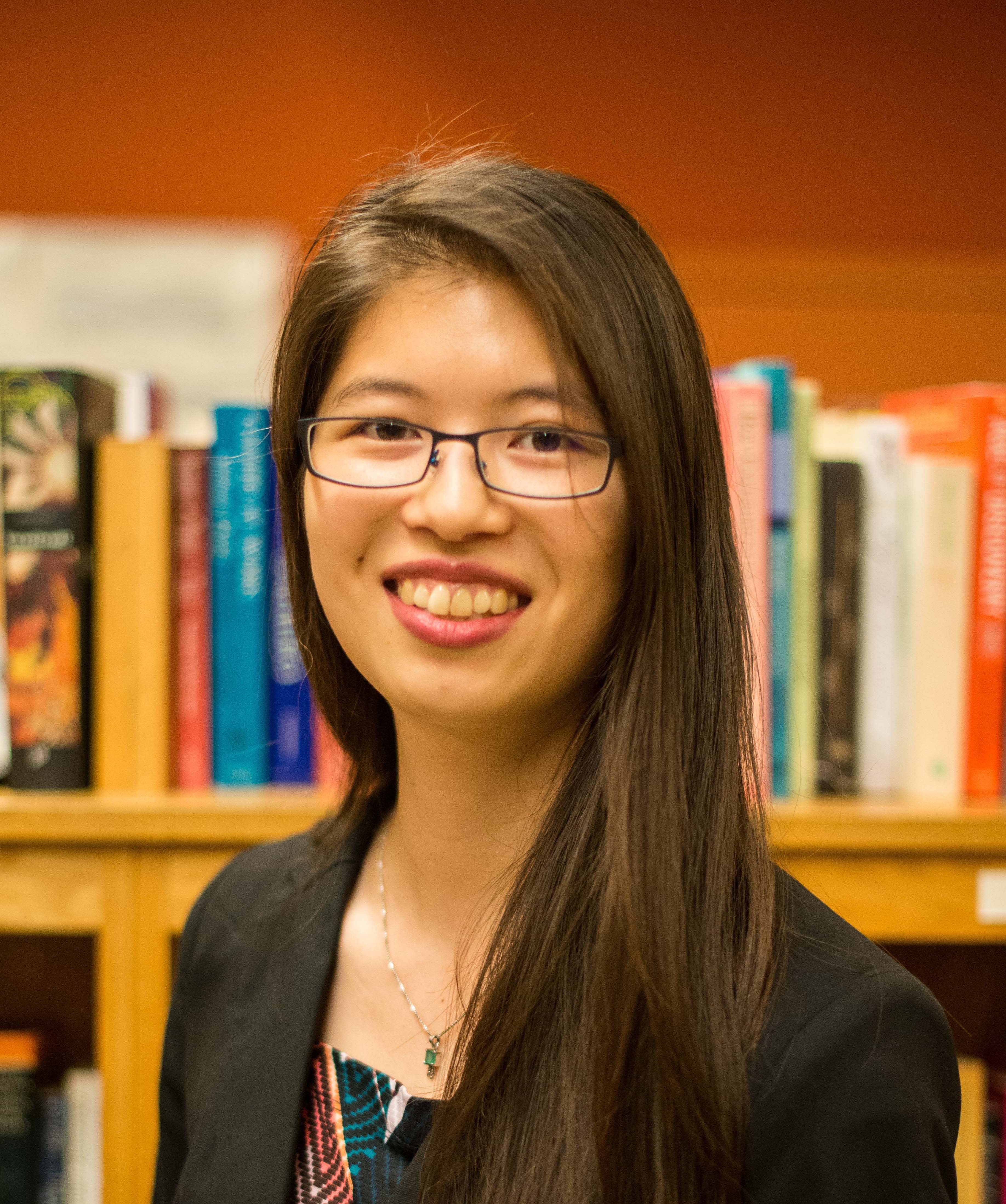}}]{Cathy Wu}
received the B.S. and M.Eng. degrees from the Massachusetts Institute of Technology (MIT), Cambridge, MA, USA, in 2012 and 2013, respectively, and the Ph.D. degree from the University of California Berkeley, Berkeley, CA, USA, in 2018, all in electrical engineering and computer sciences. She was a Postdoc with the Microsoft Research AI. She is an Assistant Professor with MIT in LIDS, CEE, and IDSS. She studies the technical challenges surrounding the integration of autonomy into societal systems. Her research interests include machine learning and mobility. Prof. Wu was a recipient of several awards, including the 2019 IEEE Intelligent Transportation Systems Society (ITSC) Best Ph.D. Dissertation Award, 2018 Milton Pikarsky Memorial Dissertation Award, and the 2016 IEEE ITSC Best Paper Award, and has appeared in the press, including Wired and Science.
\end{IEEEbiography}

%\newpage

\end{document}